\documentclass{article}
\usepackage{arxiv}

\usepackage[T1]{fontenc}
\usepackage[utf8]{inputenc}

\usepackage{amsmath,amssymb,amsfonts,amsthm}
\usepackage{orcidlink}

\usepackage{graphicx}
\usepackage{booktabs}
\usepackage{url}
\usepackage{microtype}
\usepackage{bm}
\usepackage{multirow}
\usepackage{makecell}
\usepackage[dvipsnames]{xcolor}

\usepackage{tikz}
\usetikzlibrary{trees,shapes.geometric,arrows}

\usepackage{linguex}

\usepackage{algorithm}
\usepackage[noend]{algpseudocode}

\usepackage[numbers]{natbib}

\def\b1{\mathbf{1}}

\theoremstyle{plain}

\theoremstyle{definition}

\theoremstyle{remark}
\newtheorem{remark}{Remark}

\title{How Large Language Models Get Stuck: Early structure with persistent errors}

\author{
Alokesh Manna \orcidlink{0009-0007-7958-5273} \\
Department of Statistics \\
University of Connecticut, Storrs, USA \\
\texttt{alokesh.manna@uconn.edu} \\
\And
William Snyder \orcidlink{0000-0003-1788-2057}\\
Department of Linguistics \\
University of Connecticut, Storrs, USA \\
\texttt{william.snyder@uconn.edu} \\
\And
Whitney Tabor \orcidlink{0000-0002-6818-3283}\\
Department of Psychological Sciences\\
University of Connecticut, Storrs, USA \\
\texttt{whitney.tabor@uconn.edu}
}

\begin{document}
\maketitle
\begin{abstract}

\smallskip
Large Language Models (LLMs) have achieved remarkable success in modeling natural language.   Their behavior makes it clear that they internalize a number of grammatical and semantic systematicities that linguists have identified as characteristic of natural languages.   Nevertheless, it is well known that they are expensive to train and they struggle to match some aspects of human language ability.   In this project, we ask whether insights from language theory can reveal generalizations about where LLMs succeed and where they fail, and whether these insights can lead to more effective training methods.     We have studied Meta’s OPT model trained on a BabyLM  dataset (100M words, “developmentally [more] plausible” than state-of-the-art LLMs) and evaluated it under controlled grammatical interventions using the BLiMP benchmark, which consists of 67 syntactic categories, each defined by sentence pairs that differ in a targeted grammatical rule violation.    For example, it is natural to ask, “Which paper did Jason file before he read the report?”  but it is more awkward to ask, “Which paper did Jason file the report before he read?”  This distinction falls under the heading of what linguists call “Island Constraints”---a set of generalizations about which parts of hierarchical sentence structures are available for questioning.   Using the BLiMP database,  we tested the model’s preference for grammatical over ungrammatical sentences across training iterations and grammatical types. Our results show that, in nearly one-third of the BLiMP categories, including Island Constraints, OPT fails to consistently assign a higher likelihood to grammatical sentences, even after extensive training.   When it fails, it often establishes a clear (erroneous) separation of the likelihoods at an early stage of processing when, across the board, its structural behaviors are falling into place.    We are investigating this transition point to see if it is a locus at which a difference in training strategy can improve efficiency and success.

\noindent\textbf{Keywords:} Large Language Model, metric Grammar, BabyLLM, Deep learning, change point detection, computational linguistics, Causal ML
\end{abstract}

\section{Introduction}
Large Language Models (LLMs) have made remarkable strides in capturing the grammatical and semantic regularities of natural language. Yet for all their sophistication, they remain expensive to train and fall short of human-level competence in systematic ways. Understanding where and why they fail is valuable both for the theory of neural network language learning and for making training more effective and efficient.
In this paper, we investigate whether insights from formal language theory can reveal principled generalizations about the success and failure points of LLMs. We focus on Meta's OPT model \cite{zhang_opt_2022}, that we trained on the BabyLM 100M-word corpus \cite{warstadt2023findings}---a dataset chosen for its developmental plausibility relative to the massive corpora used by state-of-the-art models---and evaluate it using the BLiMP benchmark \cite{Warstadt2020BLiMP}, a suite of 67 syntactic categories defined by minimal sentence pairs that isolate specific grammatical violations.\footnote{We chose BLiMP because it provides broad coverage of structural features of English, divided into types long recognized by syntactic researchers as central to the structural systems of natural languages generally.  It seems, at present, to be the most often-studied data base of this nature.  Nevertheless, it is weak in several regards---many sentences are semantically odd, the comparisons between ``Good" and ``Bad" sentences are not always well-controlled---see Section \ref{DiscussionSection} below, and the range of tests on each syntactic type is narrow.}
Our central finding is that the model's failures exhibit a characteristic pattern. In nearly one-third of BLiMP classes---including phenomena such as Island Constraints, NPI licensing, and binding principles---OPT consistently assigns higher likelihood to the ungrammatical member of each pair. Crucially, this erroneous preference is typically not a late-training artifact: it emerges early and stabilizes, suggesting that the model locks into an incorrect representation during a critical window of training and subsequently reinforces it. We use change-point detection methods (CUSUM and the ruptures framework) to pinpoint precisely when these divergences arise and we develop a hypothesis and a strategy for testing the hypothesis in support of characterizing the conditions under which entrenched erroneous learning occurs.

The sections are arranged as follows. First, we describe the methods of training and estimated parameters in the section~\ref{sec:methods}. Following that, we describe the results and the key major findings of the three different behaviors of the Blimp classes in the section~\ref{sec:results}. Finally, we provide a discussion of the change point for different classes of grammatical phenomena and a hypothetical justification in the line of Bigram hypotheses for why those behaviours are expected in section~\ref{DiscussionSection}, with a concluding remark in section~\ref{conclusion}. In the section~\ref{appendix}, we describe the choice of the hyperparameters and the training procedure, and additional information relevant to our experiments.

\section{Methods}
\label{sec:methods}
We followed the training protocol of \citet{zhang_opt_2022}. A brief description of the choice of the hyperparameters are given in the section~\ref{appendix}.

\textbf{Model Checkpoints and Training Trajectory.} Let $\{\theta_t\}_{t=1}^{T}$ denote the sequence of model checkpoints obtained during training, where $\theta_t$ represents the complete set of parameters of the OPT transformer after $t$ optimizer update steps. Training was conducted on the standard BabyLM 100M-word corpus using the High Performance Computing facility at $\langle$Name Omitted$\rangle$. Each checkpoint corresponds to a fixed training iteration rather than an epoch, reflecting the streaming nature of language model optimization.
Checkpoints were stored at irregular iteration indices,
$t \in \{100, 350, 1250, \ldots, 30800\}$,
with denser sampling during early training and increasingly sparse sampling at later stages. This design choice reflects a practical trade-off: early training phases exhibit rapid changes in model behavior, while storing parameters at every iteration would be computationally and storage-wise prohibitive. \\

\textbf{Perplexity and Accuracy Evaluation Across Checkpoints:}
For a causal language model (LM) with parameters $\theta$ and a tokenizer, the sentence-level negative log-likelihood of sentence $x = (w_1, \ldots, w_n)$ under the causal language model is defined as
%\footnote{Language model probability generation at any point in time is, in principle, sensitive to arbitrary preceding events. To control comparison across presentations of sentences, we set the state of the model to XXXX at the beginning of each sentence.}
\[
\mathcal{L}(x;\theta_t)
= -\frac{1}{n}\sum_{i=1}^{n} \log p_{\theta_t}(w_i \mid w_{<i}),
\]
and the corresponding perplexity is
\[
\mathrm{PPL}(x;\theta_t) = \exp\big(\mathcal{L}(x;\theta_t)\big).
\]

%%%%%% Deleted:
% For each BLiMP minimal pair $(x^{\text{good}}, x^{\text{bad}})$, checkpoint-level accuracy is defined as
% \[
% \mathbb{I}\big(\mathrm{PPL}(x^{\text{good}};\theta_t) < \mathrm{PPL}(x^{\text{bad}};\theta_t)\big),
% \]
% where $\mathbb{I}(\cdot)$ denotes the indicator function. Paradigm-level accuracy is obtained by averaging this indicator across all sentence pairs within a BLiMP category. This evaluation was performed in parallel for all $67$ BLiMP paradigms at each stored checkpoint.

For each test category, we evaluate grammatical ($s_\mathrm{good}$) and ungrammatical ($s_\mathrm{bad}$) sentence pairs. The model is considered correct on a pair if the grammatical sentence has lower perplexity:
\[
\text{Correct}_{\mathrm{pair}} =
\begin{cases}
1, & \text{if } \mathrm{PPL}(s_\mathrm{good}) < \mathrm{PPL}(s_\mathrm{bad}), \\
0, & \text{otherwise.}
\end{cases}
\]
The overall accuracy across $M (= 1000)$ sentence pairs in a category is then:
\begin{align}
\label{eqn:accuracy}
    \mathrm{Accuracy} = \frac{1}{M} \sum_{j=1}^{M} \text{Correct}_{\mathrm{pair}_j} \times 100\%.
\end{align}
For each category, we record the perplexities of the grammatical and ungrammatical sentences, and whether the model prediction was correct. These values serve as the basis for constructing the log-perplexity gap
\begin{align}  
\label{PerplexityGap}
\Delta \log \mathrm{PPL} = \log(\sum_{j=1}^{M}\mathrm{PPL}_{\mathrm{good}}) - \log(\sum_{j=1}^{M}\mathrm{PPL}_{\mathrm{bad}})
\end{align}
which is subsequently used in change-point detection via CUSUM or ruptures.

\textbf{Checkpoint-Based Change-Point Analysis.} To study the emergence of grammatical sensitivity over training,
%%%%%% Deleted:
% we define the log-perplexity gap at checkpoint $t$ as
% \[
% g_t
% = \log \mathrm{PPL}(x^{\text{bad}};\theta_t)
% - \log \mathrm{PPL}(x^{\text{good}};\theta_t),
% \]
% where positive values indicate a correct model preference for the grammatical sentence. Treating $\{g_t\}_{t=1}^{T}$ as a discrete-time sequence indexed by checkpoints, 
we apply change-point detection methods, including CUSUM and the \text{ruptures} framework, to identify iterations at which the statistical behavior of the perplexity gap ($\Delta \log \mathrm{PPL}$) shifts.  Generally, at the beginning of training, there is no perplexity gap between Good and Bad sentences. Our aim, with these measures, is to determine, by assessing the entire trajectory of the gap over the course of training, the moment when the critical separation occurs.

A detected change point corresponds to a specific checkpoint $\theta_{\hat{t}}$ and is interpreted as a training iteration at which the model undergoes a qualitative transition in its ability to distinguish grammatical from ungrammatical constructions. This checkpoint-based formulation enables direct alignment between training dynamics, accuracy trajectories, and distributional changes in perplexity across BLiMP classes.

\section{Results}
\label{sec:results}

\begin{figure*}[t]
    \centering
    \includegraphics[width=\textwidth]{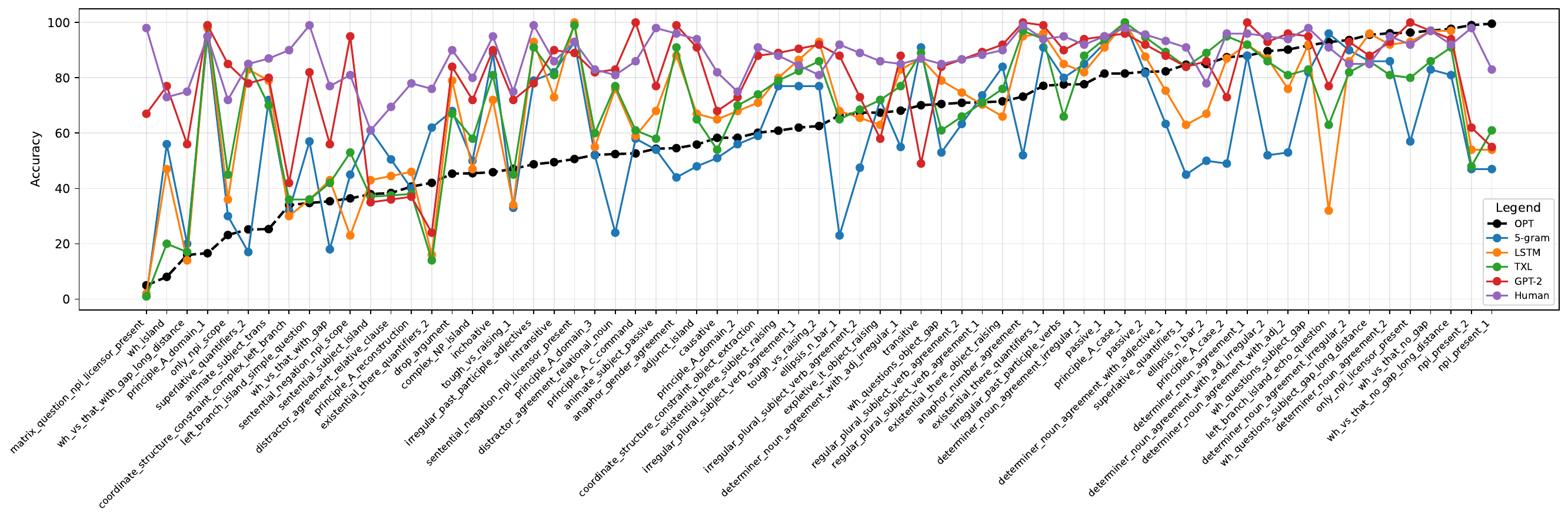}
    \caption{Accuracy (\ref{eqn:accuracy}) comparison among Blimp classes arranged by OPT performance. }
    \label{fig:accuracy-w}
\end{figure*}

\begin{figure*}[t]
    \centering
    \includegraphics[width=\textwidth]{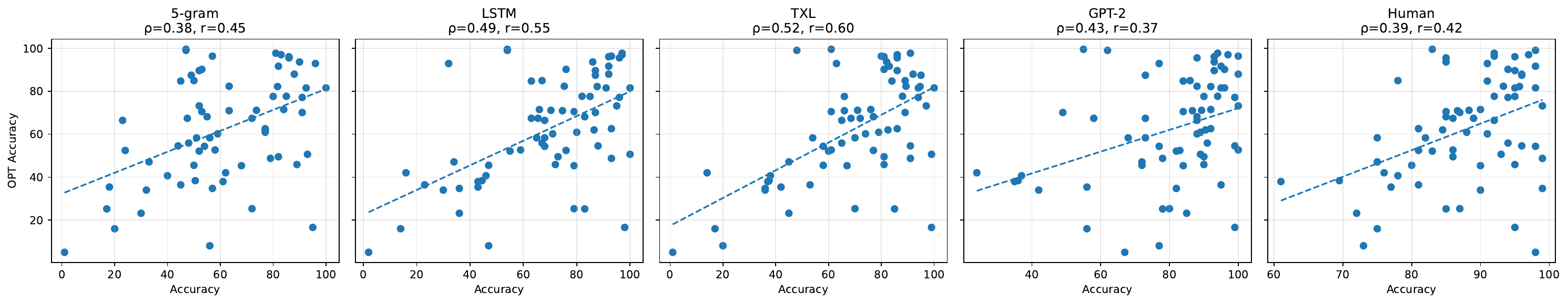}
    \caption{Correlation plot of accuracy (\ref{eqn:accuracy}) for OPT and 5 other models, which demonstrate a positive association.}
    \label{fig:accuracy-correlation}
\end{figure*}

\textbf{Accuracy-Based Stratification of BLiMP classes compared to \cite{Warstadt2020BLiMP}.} We first compared our model's end-of-training profile to that of other relevant LMs. \citet{Warstadt2020BLiMP} introduced the BLiMP dataset and provided a test of various models' performance as well as human performance across the 67 classes.   GPT-2, which was state-of-the-art at that time, performed more poorly than humans but nevertheless quite well in comparison to the humans.  The relevance of our current effort, based on training on a much smaller corpus for much less time and no fine-tuning, to the goal of improving state-of-the-art LLM performance will be more convincing if we can show that our model's performance is plausibly a precursor to the performance of GPT-2.  To this end, we compare our OPT model's accuracy—computed as the average over the final three training iterations—against the five reference models reported in Table 4 of \cite{Warstadt2020BLiMP}: a 5-gram model, LSTM, Transformer-XL (TXL), GPT-2, and humans. Figure~\ref{fig:accuracy-w} displays these comparisons with BLiMP classes ordered by ascending OPT accuracy. As expected, given our less intensive training regime, OPT achieves lower absolute accuracy than all five reference models. Nevertheless, its accuracy is positively correlated with that of all five baselines, with Pearson correlations ($\rho$) ranging from .38 to .52 and Spearman rank correlations ranging from .37 to .60 (Figure~\ref{fig:accuracy-correlation}). This convergence lends support to the view that our developmental analysis may be helpful in improving the accuracy and efficiency of more highly trained LLMs.

%%%%%% Deleted:
% To analyze the behavior of perplexity across model checkpoints, we computed the log-difference between the average perplexity on "good" and "bad" sentences for each dataset:
% \[
% \Delta \log \mathrm{PPL} = \log(\mathrm{PPL}_{good}) - \log(\mathrm{PPL}_{bad}).
% \]

\textbf{Categorization of learning trajectories.}
We now turn to categorizing the learning trajectories  for the 67 BLiMP classes. For each dataset, we first sorted the checkpoints into ascending order. Then we divided them into three segments: early (first 30\%), middle (30--70\%), and late (last 30\%). For categorization, only the early and late segments were used. 
The mean perplexity gap (\ref{PerplexityGap}) was computed for both the early and late segments:
\begin{equation}
\label{eqn:early_late}
    \bar{\mathrm{PPL}}_{j}
 = \frac{1}{n_j} \sum_{i=1}^{n_j} \Delta \log \mathrm{PPL}_i
\end{equation}
% \[
% \bar{\mathrm{PPL}}_{\text{late}}
%  = \frac{1}{n_\text{late}} \sum_{i=1}^{n_\text{late}} \Delta \log \mathrm{PPL}_i.
% \]
where $j\in\{\text{early}, \text{late}\}$ and $n_j$ is the number of checkpoints in  segment $j$.
BLiMP classes were then assigned to one of four categories based on these means:
\begin{itemize}
    \item \textbf{Erroneous early and sustained (EES)}: Blimp classes with both early and late means positive, indicating initially and consistently higher perplexity on good samples. ($\bar{\mathrm{PPL}}_{j}>0$).
    \item \textbf{Correct early and sustained (CES)}: Blimp classes with both early and late means negative, indicating initially and consistently lower perplexity on good samples. ($\bar{\mathrm{PPL}}_{j}<0$).
    \item \textbf{Correct Late Separation (CLS)}: Blimp classes with a positive early mean but a negative late mean, indicating that the separation between good and bad samples emerged only at later checkpoints. ($\bar{\mathrm{PPL}}_{\text{early}}>0, \bar{\mathrm{PPL}}_{\text{late}}<0$).
    \item \textbf{Erroneous Late Separation (ELS)}: Blimp classes with a negative early mean but a positive late mean, indicating that the separation between good and bad samples emerged only at later checkpoints. ($\bar{\mathrm{PPL}}_{\text{early}}<0, \bar{\mathrm{PPL}}_{\text{late}}>0$). 
\end{itemize}

\begin{figure*}[t]
    \centering
    \includegraphics[width=\textwidth]{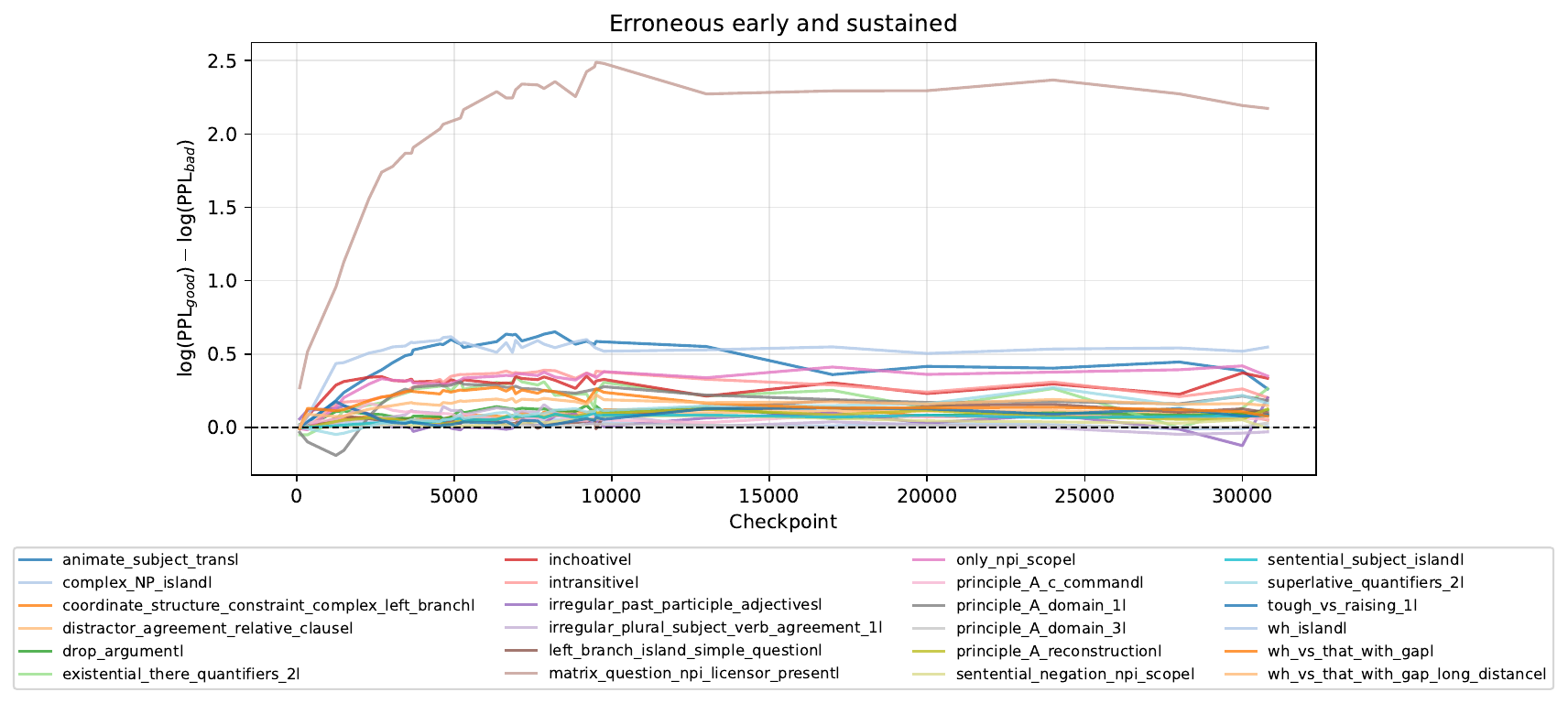}
    \caption{Erroneous early and sustained separation across checkpoints.}
    \label{fig:early_erroneous}
\end{figure*}

\begin{figure*}[t]
    \centering
    \includegraphics[width=\textwidth]{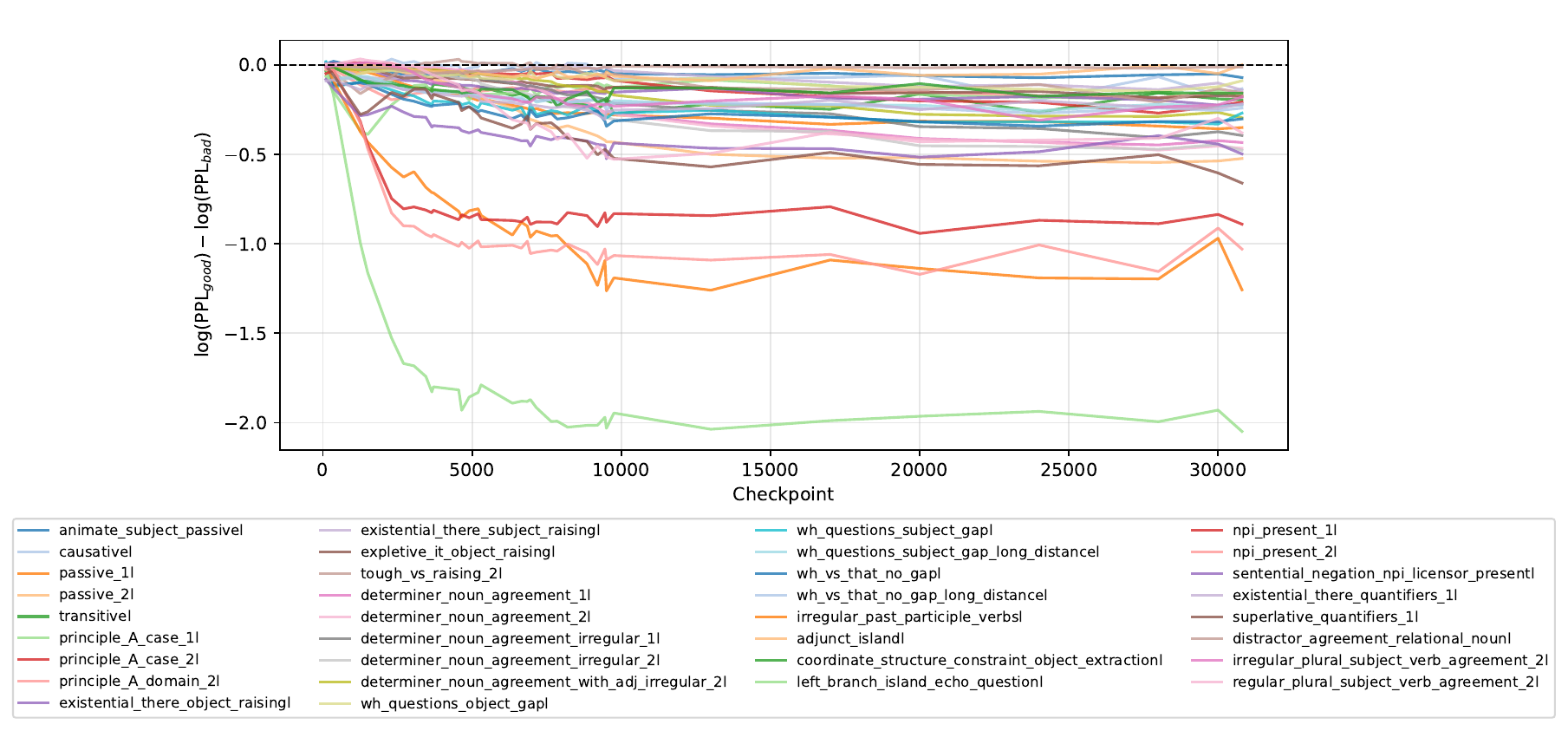}
    \caption{Correct early separation across checkpoints.}
    \label{fig:early_correct}
\end{figure*}

\begin{figure*}[t]
    \centering
    \includegraphics[width=\textwidth]{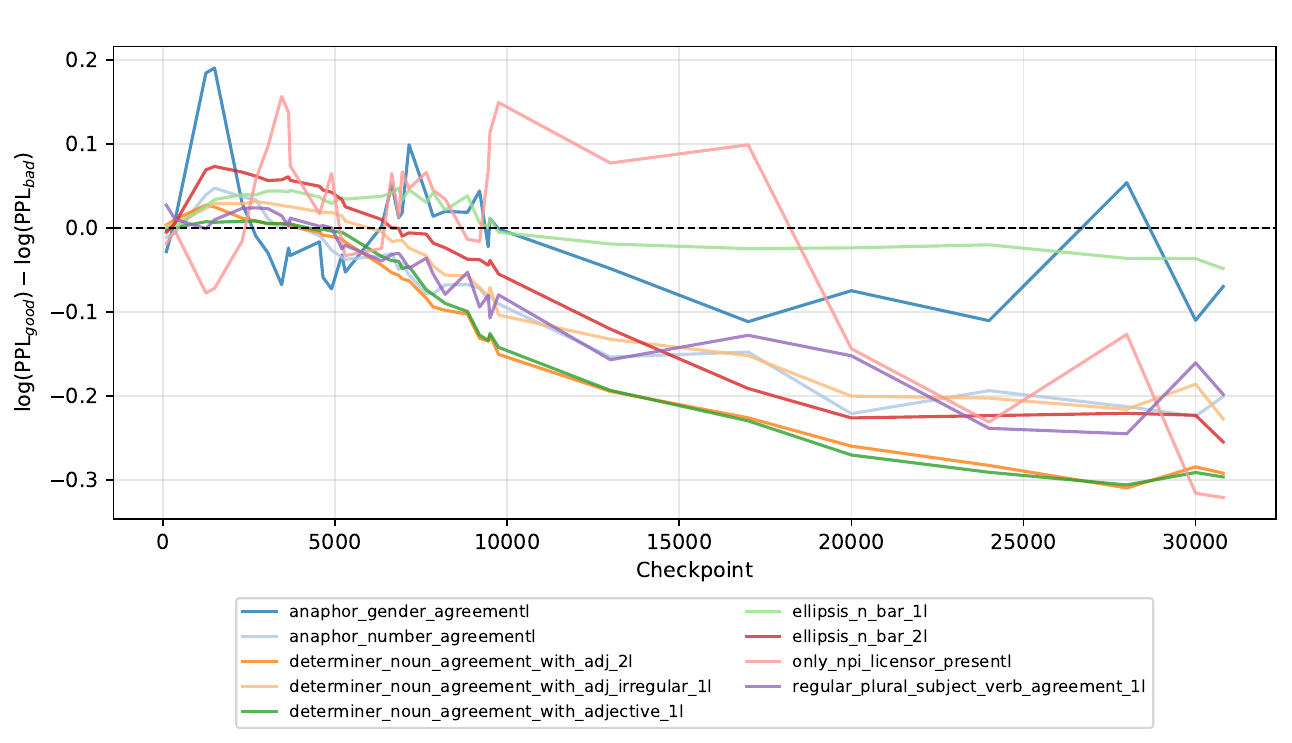}
    \caption{Late Correct Separation across checkpoints.}
    \label{fig:late_separation}
\end{figure*}

No BLiMP class fell into the ELS category, consistent with our thesis that the major structural organization is established early in training (around iteration 6000 out of 30800 total iterations).  Figures \ref{fig:early_erroneous} - \ref{fig:late_separation} show the distribution of the BLIMP classes across the three remaining categories.

\textbf{Change point detection.}
In addition to categorizing Blimp into Early/Late Correct/Erroneous Separation, we also used change-point detection methods to identify the checkpoint at which the separation between good (grammatical) and bad (ungrammatical) perplexities became statistically significant.

\begin{enumerate}

\item \textbf{CUSUM based change point detection.} ( \cite{page1955test}, \cite{hinkley1970inference}): 
To identify the checkpoint at which grammatical (good) and ungrammatical (bad) perplexities begin to separate, we applied a window-free cumulative sum (CUSUM) method. For each dataset, the checkpoints were first sorted and the perplexity gap, which we label here $\delta^p$ for conciseness, was computed at each checkpoint. For each candidate change point $t$, excluding a small fraction of checkpoints at the edges to avoid boundary effects, a CUSUM statistic was calculated as the absolute difference between the mean of the segment after $t$ and the mean of the segment before $t$:
\[
\text{CUSUM}_t = \left| \frac{1}{T-t} \sum_{i=t+1}^{T} \delta^{p}_i - \frac{1}{t} \sum_{i=1}^{t} \delta^{p}_i \right|,
\]
where $T$ is the total number of checkpoints. The checkpoint corresponding to the maximum CUSUM value is taken as the estimated change point, representing the position at which the separation between good and bad perplexities becomes most pronounced. \footnote{CUSUM is conceptually equivalent to detecting a mean shift in the gap signal without specifying a window size.}

%%%%%% Deleted (from footnote)
% The detected change point is visualized alongside the log-perplexities to facilitate interpretation.

\item \textbf{Ruptures-Based Distributional Method ((\cite{truong2018ruptures})):}  
We also used the \text{ruptures} library to detect distributional change points. Ruptures generalizes the CUSUM approach to account for variance and nonparametric shifts.

For each dataset, we computed the log-gap between bad and good perplexities, $\delta^{g} = \log(\mathrm{PPL}_{bad}) - \log(\mathrm{PPL}_{good})$, and treated it as a one-dimensional signal. The change-point problem is formulated as finding a set of breakpoints $\{\tau_1, \dots, \tau_K\}$ that partition the signal into contiguous segments such that a cost function, $C(y_{s:e})$, which measures the deviation of the segment from a model, is minimized across all segments while optionally including a penalty on the number of segments. For classical mean-shift detection, $C(y_{s:e})$ is the sum of squared deviations from the segment mean, $C(y_{s:e}) = \sum_{t=s}^{e} (y_t - \bar{y}_{s:e})^2$, whereas for a general nonparametric setting, an energy-based or kernelized cost is used, $C(y_{s:e}) = \sum_{i,j=s}^{e} K(y_i, y_j)$, where $K$ is an RBF kernel capturing differences in distribution beyond just mean and variance. The total cost is then the sum of segment costs plus any penalty, $\sum_{k=0}^{K} C(y_{\tau_k+1:\tau_{k+1}}) + \text{penalty}(K)$, and the optimal set of change points minimizes this objective. \footnote{In practice, the \text{Binseg} algorithm was used to approximate this optimization by recursively splitting the signal at the point that maximally reduces the total cost. The first detected change point corresponds to the checkpoint at which the gap between good and bad perplexities begins to separate, while datasets with no significant change are classified based on the overall mean and variance of the gap.}

The signal was trimmed at both ends to avoid edge effects, and the \text{Binseg} or \text{PELT} algorithm was applied with an energy-based (\text{rbf}) cost model. If a change point was detected, it was returned; otherwise, the dataset was classified as either \textit{always together} or \textit{always apart} based on the mean and standard deviation of the gap. Diagnostic information, including the regime, mean gap, standard deviation, and breakpoints, was recorded.

\end{enumerate}
\begin{table}[!t]
\centering
\caption{Summary of BLiMP change-point detection.}
\label{tab:blimp-summary}
\small
\begin{tabular}{lrrr}
\toprule
\textbf{Pattern} & \textbf{n} & \textbf{Avg CUSUM} & \textbf{Avg Ruptures} \\
\midrule
CES  & 34 & 7{,}351  & 6{,}234 \\
CLS                      & 9  & 20{,}367 & 6{,}994 \\
%EES       & 1  & 28{,}000 & 9{,}750 \\
EES   & 24 & 5{,}542  & 6{,}798 \\
\midrule
%\textbf{Total}                    & \textbf{67} & & \\
\bottomrule
\end{tabular}
\end{table}

Additional details for the attention layer dynamics across different training periods for query, key, vector, and output layer dynamics are available in the section~\ref{training_dynamics}. 

\section{Discussion}
\label{DiscussionSection}
\begin{table}[p]
\caption{BLiMP cases where the average perplexity over the last five checkpoints satisfies $\overline{\mathrm{PPL}}_{\mathrm{good}} < \overline{\mathrm{PPL}}_{\mathrm{bad}}$.}
\label{tab:blimp-correct-yes}
\centering
\small
%\footnotesize
\renewcommand{\arraystretch}{0.65}
\setlength{\tabcolsep}{4pt}
\begin{tabular}{lrrll}
\toprule
BLiMP case & \makecell{Change point \\ (CUSUM)} & \makecell{Change point \\ (RUPTURES)} & \makecell{Correct \\ (direction)} & Temporal pattern \\
\midrule

\multicolumn{5}{l}{\textbf{Early correct separation}} \\
\midrule
adjunct\_islandl & 2300 & 4900 & Yes & Early correct \\
animate\_subject\_passivel & 1250 & 4900 & Yes & Early correct \\
causativel & 30000 & 8200 & Yes & Early correct \\
coordinate\_structure\_constraint\_object\_extractionl & 1250 & 6850 & Yes & Early correct \\
determiner\_noun\_agreement\_1l & 20000 & 6850 & Yes & Early correct \\
determiner\_noun\_agreement\_2l & 20000 & 6850 & Yes & Early correct \\
determiner\_noun\_agreement\_irregular\_1l & 28000 & 6850 & Yes & Early correct \\
determiner\_noun\_agreement\_irregular\_2l & 20000 & 6850 & Yes & Early correct \\
determiner\_noun\_agreement\_with\_adj\_irregular\_2l & 20000 & 6850 & Yes & Early correct \\
distractor\_agreement\_relational\_nounl & 1250 & 6850 & Yes & Early correct \\
existential\_there\_object\_raisingl & 1500 & 4900 & Yes & Early correct \\
existential\_there\_quantifiers\_1l & 1250 & 4900 & Yes & Early correct \\
existential\_there\_subject\_raisingl & 20000 & 6850 & Yes & Early correct \\
expletive\_it\_object\_raisingl & 1500 & 6850 & Yes & Early correct \\
irregular\_past\_participle\_verbsl & 350 & 8200 & Yes & Early correct \\
irregular\_plural\_subject\_verb\_agreement\_2l & 3050 & 4900 & Yes & Early correct \\
left\_branch\_island\_echo\_questionl & 350 & 6850 & Yes & Early correct \\
npi\_present\_1l & 1250 & 3450 & Yes & Early correct \\
npi\_present\_2l & 1250 & 3450 & Yes & Early correct \\
passive\_1l & 1250 & 4900 & Yes & Early correct \\
passive\_2l & 6350 & 6850 & Yes & Early correct \\
principle\_A\_case\_1l & 2300 & 6850 & Yes & Early correct \\
principle\_A\_case\_2l & 28000 & 9750 & Yes & Early correct \\
principle\_A\_domain\_2l & 350 & 4900 & Yes & Early correct \\
regular\_plural\_subject\_verb\_agreement\_2l & 6350 & 4900 & Yes & Early correct \\
sentential\_negation\_npi\_licensor\_presentl & 350 & 4900 & Yes & Early correct \\
superlative\_quantifiers\_1l & 350 & 8200 & Yes & Early correct \\
tough\_vs\_raising\_2l & 28000 & 8200 & Yes & Early correct \\
transitivel & 350 & 6850 & Yes & Early correct \\
wh\_questions\_object\_gapl & 350 & 9750 & Yes & Early correct \\
wh\_questions\_subject\_gapl & 350 & 4900 & Yes & Early correct \\
wh\_questions\_subject\_gap\_long\_distancel & 350 & 4900 & Yes & Early correct \\
wh\_vs\_that\_no\_gapl & 350 & 4900 & Yes & Early correct \\
wh\_vs\_that\_no\_gap\_long\_distancel & 350 & 4900 & Yes & Early correct \\

\midrule
\multicolumn{5}{l}{\textbf{Late correct separation}} \\
\midrule
anaphor\_gender\_agreementl & 2300 & 9750 & Yes & Late separation \\
anaphor\_number\_agreementl & 20000 & 4900 & Yes & Late separation \\
determiner\_noun\_agreement\_with\_adj\_2l & 20000 & 6850 & Yes & Late separation \\
determiner\_noun\_agreement\_with\_adj\_irregular\_1l & 20000 & 6850 & Yes & Late separation \\
determiner\_noun\_agreement\_with\_adjective\_1l & 20000 & 6850 & Yes & Late separation \\
ellipsis\_n\_bar\_1l & 30000 & 8200 & Yes & Late separation \\
ellipsis\_n\_bar\_2l & 17000 & 4900 & Yes & Late separation \\
only\_npi\_licensor\_presentl & 30000 & 9750 & Yes & Late separation \\
regular\_plural\_subject\_verb\_agreement\_1l & 24000 & 4900 & Yes & Late separation \\

\midrule
\multicolumn{5}{l}{\textbf{Early erroneous and sustained}} \\
\midrule
irregular\_plural\_subject\_verb\_agreement\_1l & 28000 & 9750 & Yes & Early erroneous \\

\bottomrule
\end{tabular}
\end{table}

Tables \ref{EESTable} and \ref{CESTable} show the results of our qualitative assessment of the ESS class and the CES class.

\begin{table}[p]
\centering
\caption{Erroneous Early and Sustained (EES).  ``***Bigram" indicates that the Bigram hypothesis correctly predicts the model's ordering of average perplexities for the group.  ``\#\#\#Bigram" indicates that the Bigram hypothesis fails to predict the model's behavior with the group. \label{EESTable}}

\small
\begin{tabular}{p{7cm} p{8cm}}
\hline
\textbf{CASE} & \textbf{ANALYSIS} \\
\hline

animate\_subject\_trans & Semantic constraint \\

complex\_NP\_island & Very weak separation \\

coordinate\_structure\_constraint \_complex\_left\_branch 
& ***Bigram: High prob of “What is/has/could/should” in Bad vs “What N” in Good \\

distractor\_agreement\_relative\_clause 
& ***Bigram: Local coherence (Bad) vs Long Distance Coherence (Good) \\

drop\_argument 
& Arbitrary, distinct verb choices for Bad and Good—e.g., Good: many sentences use “buy”, “skate around”, “investigate”, “climb up”, “ascend”. 
Bad: many sentences use “listen to”, “talk to”, “escape from”, “boast about”. 
Also: some ambiguous Bad that are actually good on other readings. \\

existential\_there\_quantifiers\_2 
& Good sentences are very semantically strange. Bad are often ambiguous and fine in their other meaning. \\

inchoative 
& Arbitrary, distinct verb choices for Bad and Good. \\

intransitive 
& Arbitrary, distinct verb choices for Bad and Good.  Many Bad have verb-preposition constructions which may up their bigram likelihood (because prepositions are high-frequency words).  Relative few Good cases have verb-preposition constructions.\\

irregular\_past\_participle\_adjectives 
& Very mildly separated \\

irregular\_plural \_subject\_verb\_agreement\_1 
& Early erroneous separation, but then corrects (not sustained) \\

left\_branch\_island\_simple\_questions 
& ***Bigram: Bad with plural extracted nouns are locally felicitous (Good have long-distance dependency). 
Testing singular extracted nouns yields strongly correct behavior (detectable from local dependency). \\

matrix\_question\_npi\_licensor\_present 
& ***Bigram: Good: “Had Bruce ever …” 
Bad: “Bruce had ever” — “[Aux] ever” syntactic sticky pair \\

only\_npi\_scope 
& ***Bigram: Good: “Only NP [Comp]”, 
Bad: “NP [Comp] only” — “[Comp] only” syntactic sticky pair (terrible sentences) \\

principle\_A\_c\_command 
& Mild separation \\

principle-A\_domain\_1 
& \#\#\#Bigram: Good: Personal Pronoun 
Bad: Reflexive (Personal pronouns far more frequent as unigrams) \\

principle-A\_domain\_3 
& \#\#\#Bigram: 
Good: Female/Male V-SComp; Male/Female VTrans Reflexive\_Male/Reflexive\_Female. 
Bad: Female/Male V-SComp; Male/Female VTrans Reflexive\_Female/Reflexive\_Male 
(Matrix subject appears to control reflexive — long-distance dependency.) \\

principle\_A\_reconstruction 
& ***Bigram: Good: Object Relative Clause; 
Bad: Subject Relative Clause (latter more frequent) \\

sentential\_negation\_npi\_scope 
& Very mild separation \\

sentential\_subject\_island 
& ***Bigram: Good: “NP’s attacking Claire irritate \_.” 
Bad: “NP’s attacking \_ irritate Claire.” 
Period more frequent after NP[Obj] than after Vtrans. \\

superlative\_quantifiers\_2 
& ???Trigram: Good: The/some/every/this N Verb; 
Bad: No N Verb. Several COCA checks show Bad trigram more frequent. \\

tough\_vs\_raising\_1 
& ***Bigram: Raising verbs (Bad) very high frequency compared to tough verbs (Good) \\

wh\_island 
& ***Bigram: “revealed they/he/she” (Good) vs “revealed who” (Bad) \\

wh\_vs\_that\_with\_gap 
& ***Bigram: Good: “[Verb] who/what”, 
Bad: “[Verb] that” (Bad order of magnitude more frequent) \\

wh\_vs\_that\_with\_gap\_long\_distance 
& ***Bigram: Same situation as wh\_vs\_that\_with\_gap \\

\hline
\end{tabular}

\ \newline

{\bf Summary counts:}

\begin{tabular}{|l|l|} \hline\hline
 Type    & Count \\ \hline\hline
 +Bigram/Trigram & 12 \\
-Bigram/Trigram &   2 \\
Semantically weird &  5 \\
Barely separated & 5 \\ \hline
{\bf TOTAL:} & 24 \\ \hline
\end{tabular}

\end{table}

\begin{table}[!ht]
\centering
\caption{Correct Early and Sustained (CES). ``***Bigram" indicates that the Bigram hypothesis correctly predicts the model's ordering of average perplexities for the group.  ``\#\#\#Bigram" indicates that the Bigram hypothesis fails to predict the model's behavior with the group. \label{CESTable}}
\small
\begin{tabular}{p{7cm} p{8cm}}
\hline
\textbf{CASE} & \textbf{ANALYSIS} \\
\hline

animate\_subject\_passive 
& Separation is very slight \\

causative 
& Separation is not very early \\

passive\_1 
& ***Bigram: Good: BE + participle of transitive verb \\

passive\_2 
& ***Bigram: Same situation as passive\_1 (difference follows from lexical frequencies) \\

transitive 
& ***Bigram: Good: “V[Trans] NP”, Bad: “V[Intrans]” \\

principle\_A\_case\_1 
& ***Bigram: Good: that[Comp] + Nominative pronoun \\

principle\_A\_case\_2 
& ***Bigram: Good: “[reflexive] V[prog]”, Bad: personal pronoun counterpart \\

principle\_A\_domain\_2 
& ***Tri/4-gram: Good “[NP[gen,num]] V [[reflexive] …]” (local structural cue) \\

existential\_there\_object\_raising 
& ***Bigram: Good: “[V[Subj-Raising]] there” \\

existential\_there\_subject\_raising 
& ***Bigram: Good: “(there) is [V[-Subj-Raising]]” \\

expletive\_it\_object\_raising 
& ***Bigram: Good: “[V[Obj-Raising]] it to be” \\

tough\_vs\_raising\_2 
& ***Bigram: Good: “BE [Subj-Raising]”, Bad: “BE [Tough]” (frequency contrast) \\

determiner\_noun\_agreement\_1 
& ***Bigram: Good: “Det[Sg/Pl] N[Sg/Pl]”, Bad: mismatched number \\

determiner\_noun\_agreement\_2 
& ***Bigram: Good: “Det[Sg/Pl] N[Sg/Pl]”, Bad: mismatched number \\

determiner\_noun\_agreement\_irregular\_1 
& ***Bigram: Same as determiner\_noun\_agreement\_1 \\

determiner\_noun\_agreement\_irregular\_2 
& ***Bigram: Same as determiner\_noun\_agreement\_1 \\

Determiner\_noun\_agreement\_with\_adjective\_irregular\_1 
& ***Trigram: Same as determiner\_noun\_agreement patterns (with adjective) \\

wh\_questions\_object\_gap 
& ***Unigram/Bigram: Good: “Noun that”, Bad: wh-form contrast \\

wh\_questions\_subject\_gap 
& ***Unigram/Bigram: Same as wh\_questions\_object\_gap \\

wh\_questions\_subject\_gap\_long\_distance 
& ***Unigram/Bigram: Same as wh\_questions\_object\_gap \\

wh\_vs\_that\_no\_gap 
& ***Bigram: Good: “Verb that”, Bad: “Verb who/which” \\

wh\_vs\_that\_no\_gap\_long\_distance 
& ***Bigram: Good: “Verb that”, Bad: “Verb who/which” \\

irregular\_past\_participle\_verbs 
& ***Bigram: Good: “NP V[tensed]”, Bad: “NP V[participle]” \\

adjunct\_island 
& Not well separated \\

coordinate\_structure\_constraint\_object\_extraction 
& ***Bigram: Good: “Noun and”, Bad: “V[trans] and” \\

left\_branch\_island\_echo\_question 
& ***Bigram: Good: “which N”, Bad: “V[trans] N[…]” \\

npi\_present\_1 
& ***Bigram: Good: either “N Adv[high freq]” or licensed NPI context \\

npi\_present\_2 
& ***Bigram: Same as npi\_present\_1 \\

sentential\_negation\_npi\_licensor\_present 
& ***Bigram: Good: “not ever”, Bad: “really/probably ever” \\

existential\_there\_quantifiers\_1 
& ***Bigram/Trigram: Good: “there BE Q[Exist]” \\

superlative\_quantifiers\_1 
& ***Bigram: Good: “fewer/more than”, Bad: “at least” pattern contrast \\

distractor\_agreement\_relational\_noun 
& Very barely separated \\

irregular\_plural\_subject\_verb\_agreement\_2 
& ***Bigram: Good: “N[Sg/Pl] V[Sg/Pl]”, Bad: mismatched number \\

regular\_plural\_subject\_verb\_agreement\_2 
& ***Bigram: Same as irregular\_plural\_subject\_verb\_agreement\_2 \\
\hline
\end{tabular}

\ \newline

{\bf Summary counts:}

\begin{tabular}{|l|l|} \hline\hline
 Type    & Count \\ \hline\hline
 +Bigram/Trigram & 30 \\
-Bigram/Trigram &   0 \\
Semantically weird &  0 \\
Barely separated & 4 \\ \hline
{\bf TOTAL:} & 34 \\ \hline
\end{tabular}
\end{table}

\begin{table}[!h]
\caption{BLiMP paradigms where the average perplexity over the last five checkpoints does not satisfy $\overline{\mathrm{PPL}}_{\mathrm{good}} < \overline{\mathrm{PPL}}_{\mathrm{bad}}$, grouped by temporal separation pattern.}
\label{tab:blimp-correct-no}
\centering
\small
%\footnotesize
\renewcommand{\arraystretch}{0.65} % <-- reduce row height
\setlength{\tabcolsep}{6pt}
\begin{tabular}{p{6cm}rrc}
\toprule
BLiMP case &
\makecell{Change point \\ (CUSUM)} &
\makecell{Change point \\ (Ruptures)} &
\makecell{Correct \\ (direction)} \\
\midrule
\multicolumn{4}{l}{\textbf{Early Erroneous Separation}} \\
\midrule
animate\_subject\_transitive & 1250 & 3450 & No \\
complex\_NP\_island & 28000 & 9750 & No \\
coordinate\_structure\_constraint\_complex\_left\_branch & 1250 & 6850 & No \\
distractor\_agreement\_relative\_clause & 2300 & 4900 & No \\
drop\_argument & 1250 & 4900 & No \\
existential\_there\_quantifiers\_2 & 1250 & 3450 & No \\
inchoative & 350 & 9750 & No \\
intransitive & 1250 & 3450 & No \\
irregular\_past\_participle\_adjectives & 2300 & 6850 & No \\
left\_branch\_island\_simple\_question & 13000 & 9750 & No \\
matrix\_question\_npi\_licensor\_present & 350 & 4900 & No \\
only\_npi\_scope & 1250 & 4900 & No \\
principle\_A\_c\_command & 350 & 4900 & No \\
principle\_A\_domain\_1 & 2300 & 9750 & No \\
principle\_A\_domain\_3 & 350 & 6850 & No \\
principle\_A\_reconstruction & 1250 & 6850 & No \\
sentential\_negation\_npi\_scope & 7650 & 8200 & No \\
sentential\_subject\_island & 1250 & 4900 & No \\
superlative\_quantifiers\_2 & 24000 & 4900 & No \\
tough\_vs\_raising\_1 & 13000 & 9750 & No \\
wh\_island & 350 & 9750 & No \\
wh\_vs\_that\_with\_gap & 350 & 9750 & No \\
wh\_vs\_that\_with\_gap\_long\_distance & 350 & 4900 & No \\
\bottomrule
\end{tabular}
\end{table}
\textbf{Change point inference:}
Applying the categorization scheme described in Section~\ref{sec:methods} to all 67 BLiMP classes yielded three empirically attested temporal patterns: \textbf{Correct Early Separation} (CES) (33 cases, $\bar{\mathrm{PPL}}_{\text{early}} < 0$, $\bar{\mathrm{PPL}}_{\text{late}} < 0$), \textbf{Erroneous Early and Sustained} (EES) (24 cases, $\bar{\mathrm{PPL}}_{\text{early}} > 0$, $\bar{\mathrm{PPL}}_{\text{late}} > 0$), and \textbf{Correct Late Separation} (CLS) (9 cases, $\bar{\mathrm{PPL}}_{\text{early}} > 0$, $\bar{\mathrm{PPL}}_{\text{late}} < 0$). The fourth logically possible pattern ($\bar{\mathrm{PPL}}_{\text{early}} < 0$, $\bar{\mathrm{PPL}}_{\text{late}} > 0$) was not observed. Classification is based on the sign of $\bar{\mathrm{PPL}}_{j}$ in the early and late segments rather than on raw accuracy alone, because $\Delta \log \mathrm{PPL}$ captures preference magnitude continuously, whereas accuracy alone may mask small, noisy, or incorrectly trending values of $\bar{\mathrm{PPL}}_{j}$ early in training. \textsc{cusum} and ruptures were then applied to identify the precise checkpoint $\hat{t}$ at which $\Delta \log \mathrm{PPL}$ shifts, providing a statistically grounded transition estimate. 

For the \textbf{CUSUM change points}, a one-way ANOVA revealed a significant effect of category on checkpoint timing ($F(2, 64) = 8.25$, $p = 0.0006$). Pairwise comparisons showed that \emph{Correct Early and sustained} paradigms reached separation significantly earlier than \emph{Correct Late Seperation} paradigms ($t = -4.00$, $p = 0.0011$), whereas there was no significant difference between \emph{Correct Early and sustained} and \emph{Erroneous Early and sustained} paradigms ($t = 0.71$, $p = 0.48$). Additionally, \emph{Correct Late Separation} paradigms separated significantly later than \emph{Erroneous Early and sustained} paradigms ($t = 4.52$, $p = 0.0004$). These results were supported by a nonparametric Kruskal--Wallis test, which confirmed overall group differences ($H = 11.89$, $p = 0.0026$). Together, these findings are consistent with our thesis that there is a critical structure-formation phase that culminates around iterations 5000-7000 and solidifies many correct and also a nontrivial number of incorrect grammatical distinctions.

CUSUM provided clearer differentiation across learning categories and captured both early and late transitions, whereas Ruptures produced more conservative, overlapping estimates. For this reason, CUSUM was selected as the primary method for reporting change‑point results.

%%%%%%% Deleted (redundant with statement below):
% This is particularly consequential for the erroneous early and sustained cases, where $\bar{\mathrm{PPL}}_{\text{early}} > 0$ is established before iteration $t = 2{,}000$ and never corrects, as detailed in Tables~\ref{tab:blimp-correct-yes} and~\ref{tab:blimp-correct-no}.

\textbf{Methodological comparison with \citet{bunzeck2024fifty}.}
Both \citet{bunzeck2024fifty} and the present study use the BLiMP benchmark \citep{warstadt_blimp_2020}
to investigate syntactic learning dynamics in language models. Both arrive at the
observation that a substantial subset of BLiMP classes -- particularly those involving island
constraints, NPI licensing, and binding -- prove persistently difficult across model families and
training regimes. Where the two approaches diverge is in their analytical goals and methods.
\citet{bunzeck2024fifty} offer a broad descriptive account, cataloging the diversity of learning curve
shapes (power-law, S-shaped, U-shaped, and various ill-behaved variants) across ten models from
two architectural families, using qualitative shape classification aided by polynomial fitting.
The present study instead pursues a causal and interventionist account: rather than characterizing
\emph{what} curves look like, we ask \emph{when} and \emph{why} failures are established.  Our analysis raises the possibility that some of the complex curve structures observed by \citet{bunzeck2024fifty} at later stages of training, might reflect the model's effort to recover from early erroneous separations established in the formative interval on which our analysis focuses. \newline

\begin{remark}
    Recall that, for a causal language model (LM) with parameters $\theta$ and a tokenizer, the sentence-level negative log-likelihood of a sentence
\[
x = (w_1, \ldots, w_n)
\]
is defined as
\[
\mathcal{L}_\theta(x)
\;=\;
-\sum_{i=1}^{n} \log p_\theta\!\left(w_i \mid w_{<i}\right),
\]
where $w_{<i} = (w_1, \ldots, w_{i-1})$.

The \emph{Bigram Hypothesis} posits that the model’s predictive distribution effectively depends only on the immediately preceding token. Formally,
\[
p_\theta\!\left(w_i \mid w_{<i}\right)
\;\approx\;
p_\theta\!\left(w_i \mid w_{i-1}\right),
\quad i \ge 2,
\]
with $w_0 = \langle \mathrm{BOS} \rangle$ denoting a fixed start-of-sequence symbol.

Under this hypothesis, the sentence-level negative log-likelihood can be approximated by the bigram loss
\[
\mathcal{L}_\theta^{(\mathrm{bi})}(x)
\;=\;
-\sum_{i=1}^{n} \log p_\theta\!\left(w_i \mid w_{i-1}\right).
\]
\end{remark}

\textbf{The Bigram Hypothesis.}
We would like to know why some categories exhibit early erroneous separation that is sustained over the course of training (EES) and some do not (CES).  The Bigram Hypothesis is our proposed explanation for this pattern. 

Focusing first on an EES case, the idea is that a specific set of conditions causes the model to be pulled strongly in the wrong direction at an early point in the training process, leading to an erroneous encoding structure.  The only pressure to counter this error comes weak and late, so it is expensive---i.e., it requires many training iterations---to overcome. For example, one of the cases yielding an EES pattern is BLiMP's Tough-vs-Raising class, which tests contrasts like that in \ref{BlimpTough}.

\ex. \label{BlimpTough}
(a) Patrick is irritating to talk to.
 (``Good") \\
(b) Patrick is about to talk to. (``Bad")
%%%%% Note:  actual (lousy) example from BLiMP uses "respect" in place of
%%%%% "talk to" (though "talk to" is used in many other test cases in
%%%%% this set.
Here the model needs to associate the choice of a 'tough' predicate (``irritating") versus a raising  predicate (``about") at an early point in the sentence, with the presence versus absence of a "gap" (e.g., talk to \_\_) at a later point. Ideally, there should be much greater perplexity at the end of (2b) than (2a).  Yet, there is another potential source of perplexity difference which is more local (involving adjacent words) and which goes in the opposite direction (penalizing 2a more than 2b): After a copula (``is"), the occurrence of the 'tough' predicate ``irritating" is far less likely than the occurrence of the raising predicate ``about". Based on COCA Corpus statistics \cite{davies_corpus_2008}\footnote{COCA = Corpus of Contemporary American English.} the frequency ratio between these two events is 1:649 (favoring ``about"). The rate of occurrence of ``to" after ``irritating" is actually GREATER than the rate of ``to" after ``about", but the asymmetry is very mild in this case (3:1 in COCA). This makes it so that, if perplexity were to be calculated solely as the average perplexity across adjacent-word pairs (i.e., ``bigrams"), the average for \ref{BlimpTough}b (96.1) would be much lower than the average for \ref{BlimpTough}a (1351.2), giving rise to an erroneous model judgment about the relative felicity of these two examples. This is just one case, but we will argue below that this pattern applies to most of the examples in BLiMP Tough-vs-Raising.

In fact, there is reason to believe that, early in training, OPT {\it does}, in fact, behave approximately like a bigram model. 
An N-gram statistical model built from a language corpus  is a model that estimates the  probability of each next-word in the corpus, by determining at what rate that word occurs after the N-1 preceding words \cite{shannon_mathematical_1948}.\footnote{If the N-1 preceding words do not occur in the corpus at all, then ``backing off" methods of estimating this probability with other data points are applied.}  It is well known that Artificial Neural Networks trained by gradient descent learning (as is our OPT model) struggle more with longer than shorter distance dependencies.  This is true both of recurrent architectures \cite{pascanu_difficulty_2013} and transformer architectures \cite{lakretz_can_2022,zimerman_long_2023} like OPT.   Based on these observations we suggest that, approximately speaking, such a network first learns a 1-gram statistical model, then a 2-gram (``bigram") model, then a 3-gram model etc.  This is only an approximation:  (i) the network will pick up on category structure relatively early and use this, in effect, to handle missing or weak N-gram data; (ii) as N grows larger, given that linguistic structures are not organized around phrase lengths specifically, the model’s behavior deviates from this simple portrayal. But early on, we claim it's a reasonable approximation.  If so, there is a stage where the model behaves approximately like a bigram model.  Our ``Bigram Hypothesis” says that if, at this stage, the bigram statistics steer the model strongly in the wrong direction with respect to a {BLiMP} minimal pair distinction, then it will get stuck in wrong expectations about this group, giving rise to the persistent erroneous perplexity separations that we have observed.  

As a step toward assessing this hypothesis, we made a qualitative survey of the BLiMP classes that fall into the two categories EES and CES.  We will describe our qualitative approach momentarily, but first, we indicate where this is leading:  Of 14 EES groups that we qualitatively judged to provide useful tests of the Bigram Hypothesis and qualitatively assessed via arguments like that presented above, we judged 12 to be aligned with the hypothesis (EES total N = 24).  Of 30 CES groups that we qualitatively found to be appropriate tests, we judged all 30 to be aligned with our hypothesis (CES total N = 34).   The number of qualifying EES cases is thus arguably large enough that addressing the model's challenges in this area could be both analytically revealing and practically useful.  Therefore, we plan, in time for the conference, to implement a bigram model (building the model from the same 100M word BabyLM data that we used to train the OPT model) and to test the Bigram Hypothesis systematically on the BLiMP classes, holding it to the standard that it should correctly predict our OPT model's separation behavior in all cases that we have not disqualified through the qualitative analysis. It is partly this disqualification possibility that makes the qualitative analysis important.

We now describe our qualitative method. We first omitted consideration of 9 BLiMP classes for which the perplexities failed to separate strongly because it is hard to qualitatively judge frequency relationships with enough precision to make predictions about these patterns.  Then, we ruled out a further 5 cases based on qualitative arguments that the comparison of Good and Bad cases was not well controlled.  An example of such a case is the Intransitive class, which is in the category EES.  The Intransitive class is meant to help determine whether the model has learned the distinction between transitive and intransitive verbs.  An example of a pair of sentences from this class is given in \ref{BlimpIntrans}.

\ex. \label{BlimpIntrans} 
(a) Robert has saluted. (``Good") \\
(b) Robert has gone to. (``Bad")

Considering the bigram model over just the sections of these sentences for which the bigrams differ (namely, ``has saluted ." vs ``has gone to ."), the perplexity, under the bigram model, of (\ref{BlimpIntrans}a) is 5112.0 while the perplexity of (\ref{BlimpIntrans}b) is 83.8 (COCA counts).  Two reasons for this big (erroneously signed) perplexity gap is that ``go" is a much more common predicate than ``salute" and ``go to" is a very common locution with ``go".  Moreover, many of the examples in the BLiMP Intransitive class exhibit a similar asymmetry.  For example, the intransitive predicates in the class are all relatively low-frequency single-word intransitive verbs (e.g., ``yawn", ``protest", ``complain", ``vaporize", ``scream", ``cook").  The transitives, by contrast include many compound (Verb + Preposition) expressions (e.g., ``go to", ``walk to", ``boast about", ``argue about", ``drive to", ``look like").  Prepositions are very high frequency words, so their inclusion in the transitive examples greatly reduces the bigram perplexity of this class. This is an arbitrary difference between the examples chosen for the test---there is no particular reason that the intransitive conditions must be low frequency and the transitive must include very high frequency words. Thus the EES classification is probably misleading. Pending confirmation of this estimate by a thorough quantitative assessment, we consider the learning trajectory of this {BLiMP} class not informative with respect to the question of how to improve LM learning.

By contrast, the {BLiMP} class Tough-vs-Raising compares examples like those shown above in \ref{BlimpTough}.  As we noted with that example, the frequency contrast between ``irritating" (Good) and ``about" (Bad) penalizes the Good and rewards the Bad.  One might argue that this is also an arbitrary difference between the particular words chosen for the examples.  But, in this case, the difference is not arbitrary. Raising verbs are among the most abstract predicates in the language---many can be used with any embedded proposition---and they are, in consequence, the most frequent of all verbs.  `Tough' verbs are also very abstract, but they impose semantic constraints on their embedded predicates (e.g, the embedded subject must be sentient), so they are relevant in fewer situations and occur less frequently overall. This is not an incidental difference but a structural difference stemming from the semantics of two kinds of verbs.  Therefore, for the Tough-vs-Raising BLiMP test, the challenging Bigram distraction is a true problem that the learning process needs to overcome.  We therefore count this case as a legitimate target of our interest in helping the model avoid or overcome early erroneous prejudices.

\section{Conclusions}
\label{conclusion}
We trained an OPT LM on the 100M word BabyLM dataset and studied its developmental performance on the BLiMP benchmark.  Our main finding was that there was a large subset of BLiMP classes that developed an erroneous average classification of the test items early in the course of training, and failed to overturn that inconsistency through the remainder of our training process.  There was also an even larger set that set off on a correct footing and sustained that.  We suggested that this early sorting of the behaviors was a global structural organization process that strongly shaped the network's organizational scheme.  We furthermore suggested that a possible route toward designing more efficient and effective training schemes could be to focus effort on steering the network more fully onto the correct path with respect to deep linguistic structural distinctions at this early stage.  In support of this effort, we introduced a hypothesis, the Bigram Hypothesis, which holds that early erroneous behavior is driven by the presence of misleading bigram statistics that dominate the network's organizational process at this stage.   We additionally described a qualitative method of assessing the value of each BLiMP test, and used this method to distinguish useful from non-useful BLiMP tests with regard to the goal of detecting structural waywardness in the critical formation phase.  Finally, we described a process we are currently implementing of quantitatively assessing the Bigram Hypothesis. We believe our approach offers a novel, and potentially useful way of blending traditional linguistic analysis with computational exploration to gain greater insight into how LLMs learn.

The description and qualitative assessment of the bigram hypotheses is presented in the Tables \ref{EESTable} and \ref{CESTable}.

\section{Funding} Alokesh Manna was supported for the Machine Learning Position by the Institute of Brain and Cognitive Sciences (\url{https://braincognitivesciences.institute.uconn.edu/}) for Spring 2025 and Fall 2026.

\section{Acknowledgments}
We sincerely thank the Institute of Brain and Cognitive Science and the College of Liberal Arts and Sciences at the University of Connecticut for establishing the Machine Learning position and for supporting this project. We are especially grateful to Professor Ofer Harel (\url{https://ofer-harel.uconn.edu/}) for initiating this position. We also thank the University of Connecticut High Performance Computing team for providing the computational resources used in this work.

% \section{Copyrights}

% The Language Resources and Evaluation Conference (LREC) Proceedings are published by the European Language Resources Association (ELRA). They are available online from the conference website.

% ELRA's policy is to acquire copyright for all LREC contributions. In assigning your copyright, you are not forfeiting your right to use your contribution elsewhere. This you may do without seeking permission and is subject only to normal acknowledgment to the LREC proceedings. The LREC Proceedings are licensed under CC-BY-NC, the Creative Commons Attribution-Non-Commercial 4.0 International License.
%\section{References}

\newpage
\bibliographystyle{apalike}
\bibliography{ref}

\section{Appendix}
\label{appendix}
\subsection{Overview}

The training procedure follows a \textbf{staged checkpoint-capture strategy}, wherein a single
base training configuration is executed repeatedly with incrementally increasing
\texttt{save\_steps} values. This allows fine-grained analysis of model learning dynamics
across different training horizons.

% -----------------------------------------------------------
\subsection{Hyperparameters}

\begin{center}
\begin{tabular}{ll}
\toprule
\textbf{Parameter} & \textbf{Value} \\
\midrule
Architecture        & OPT (decoder-only Transformer) \\
Hidden size         & 768 \\
Attention heads     & 12 \\
Hidden layers       & 12 \\
Vocabulary size     & 16{,}384 \\
Block (context) size & 256 tokens \\
\midrule
Batch size (per device) & 32 \\
Gradient accumulation steps & 4 \\
Effective batch size & 128 \\
Epochs              & 20 \\
\midrule
Optimizer           & AdamW \\
$\beta_1$, $\beta_2$ & 0.9,\ 0.98 \\
$\varepsilon$       & $10^{-6}$ \\
Weight decay        & 0.1 \\
Max gradient norm   & 1.0 \\
\midrule
Learning rate       & $2\times10^{-4}$ \\
LR scheduler        & Cosine \\
Warmup steps        & 32{,}000 \\
\midrule
Precision           & FP16 mixed \\
Seed                & 42 \\
\bottomrule
\end{tabular}
\end{center}

% -----------------------------------------------------------
\subsection*{Algorithm}

\begin{algorithm}[H]
\caption{BabyLM OPT Staged Checkpoint Training}
\begin{algorithmic}[1]

\Require Training corpus $\mathcal{D}$, max lines $N = 10^8$, output directory $\mathcal{O}$
\Require Model config $\theta = \{\text{layers}=12,\ \text{heads}=12,\ \text{vocab}=16384,\ d=768\}$

% ---- Environment ----
\State \textbf{Setup:} Initialize Conda environment; redirect HuggingFace caches to project space.

% ---- Data ----
\State \textbf{Data preparation:}
\State \quad $\mathcal{D}_{\text{train}} \leftarrow \textsc{Head}(\mathcal{D},\ N)$ \Comment{Extract first $N$ lines as training chunk}
\State \quad $\mathcal{D}_{\text{val}} \leftarrow$ BabyLM test split

% ---- Untrained baseline ----
\State \textbf{Baseline capture} (Step 0):
\State \quad Initialize model with random weights $\theta_0$
\State \quad Evaluate $\mathcal{L}_{\text{val}}(\theta_0)$ without any gradient updates \Comment{Untrained perplexity baseline}
\State \quad Save checkpoint $\theta_0 \rightarrow \mathcal{O}/\text{step\_0}$

% ---- Fine-grained phase ----
\State \textbf{Fine-grained checkpoint phase:}
\For{$s \in \{50,\ 100,\ 150,\ \ldots,\ 1000\}$} \Comment{Step $= 50$ increment}
    \State Train OPT$_\theta$ on $\mathcal{D}_{\text{train}}$ for 20 epochs using AdamW + cosine LR
    \State Save checkpoint every $s$ optimizer steps $\rightarrow \mathcal{O}/\text{step\_}s$
    \State Evaluate $\mathcal{L}_{\text{val}}(\theta)$ at each saved checkpoint
\EndFor

% ---- Coarse phase ----
\State \textbf{Coarse checkpoint phase:}
\For{$s \in \{1500,\ 2000,\ 2500,\ \ldots,\ 5000\}$} \Comment{Step $= 500$ increment}
    \State Train OPT$_\theta$ on $\mathcal{D}_{\text{train}}$ for 20 epochs using AdamW + cosine LR
    \State Save checkpoint every $s$ optimizer steps $\rightarrow \mathcal{O}/\text{step\_}s$
    \State Evaluate $\mathcal{L}_{\text{val}}(\theta)$ at each saved checkpoint
\EndFor

\State \textbf{Cleanup:} Deactivate Conda environment.

\end{algorithmic}
\end{algorithm}

% -----------------------------------------------------------
\subsection{Rationale}

The two-phase checkpoint schedule reflects a deliberate design choice:

\begin{itemize}
    \item \textbf{Fine-grained phase} ($s = 50 \to 1000$, step 50): Captures rapid early-learning
          dynamics, where loss decreases most steeply. Dense snapshots allow analysis of
          loss curves, gradient norms, and emergent linguistic behaviours at a granular level.

    \item \textbf{Coarse phase} ($s = 1500 \to 5000$, step 500): Covers later training where
          incremental improvements slow down. Sparser checkpoints reduce storage overhead
          while still tracking convergence behaviour and potential overfitting.

    \item \textbf{Untrained baseline} (Step 0): Provides a reference point for computing
          learning gains and verifying that training is proceeding correctly from random
          initialisation.
\end{itemize}

\subsection{Attention Projections and Layerwise Training Dynamics:}
\label{training_dynamics}
Transformer models (\cite{vaswani_attention_2017}) compute contextual representations using the
scaled dot-product attention mechanism.
Given a sequence of hidden states
\(
H = [h_1,\dots,h_n] \in \mathbb{R}^{n\times d}
\),
three linear projections produce the query, key, and value matrices

\[
Q = HW_Q, \qquad
K = HW_K, \qquad
V = HW_V,
\]

where

\[
W_Q, W_K, W_V \in \mathbb{R}^{d\times d}.
\]

The attention function is

\[
\mathrm{Attention}(Q,K,V)
=
\mathrm{softmax}
\left(
\frac{QK^{\top}}{\sqrt{d_k}}
\right)V ,
\]

where \(d_k\) is the key dimension.
The scaling factor \(1/\sqrt{d_k}\) prevents the dot products from
growing too large as the dimension increases.

Modern transformers use \emph{multi-head attention}.
Instead of computing a single attention operation,
the projections are split into \(h\) heads

\[
Q_i = HW_Q^{(i)}, \quad
K_i = HW_K^{(i)}, \quad
V_i = HW_V^{(i)}, \qquad i=1,\dots,h,
\]

and attention is computed independently for each head:

\[
\text{head}_i
=
\mathrm{Attention}(Q_i,K_i,V_i).
\]

The outputs are concatenated and projected back to the model space

\[
Z = (\text{head}_1,\dots,\text{head}_h),
\qquad
\mathrm{MHA}(H)=ZW_O,
\]

where \(W_O\in\mathbb{R}^{d\times d}\).
In the OPT-125M architecture used in this study,
\(d=768\) and the model uses \(h=12\) attention heads,
each operating on subspaces of dimension \(d_k=d/h=64\).

To study how these attention projections evolve during training,
we track statistics of the projection matrices
\(W_Q,W_K,W_V,W_O\) across checkpoints.
For any matrix \(W\in\mathbb{R}^{d\times d}\),
the Frobenius norm

\[
\|W\|_F =
\sqrt{\sum_{i=1}^{d}\sum_{j=1}^{d} W_{ij}^2}
\]

measures the overall magnitude of the parameters,
while the mean and standard deviation of the entries capture
their central tendency and dispersion.

The mean and standard deviation of the matrix entries are

\[
\mathrm{mean}(W) =
\frac{1}{d^2}
\sum_{i=1}^{d}\sum_{j=1}^{d} W_{ij},
\]

\[
\mathrm{std}(W) =
\sqrt{
\frac{1}{d^2}
\sum_{i=1}^{d}\sum_{j=1}^{d}
\left(W_{ij}-\mathrm{mean}(W)\right)^2 }.
\]

\begin{figure*}[t]
    \centering
    \includegraphics[width=\textwidth]{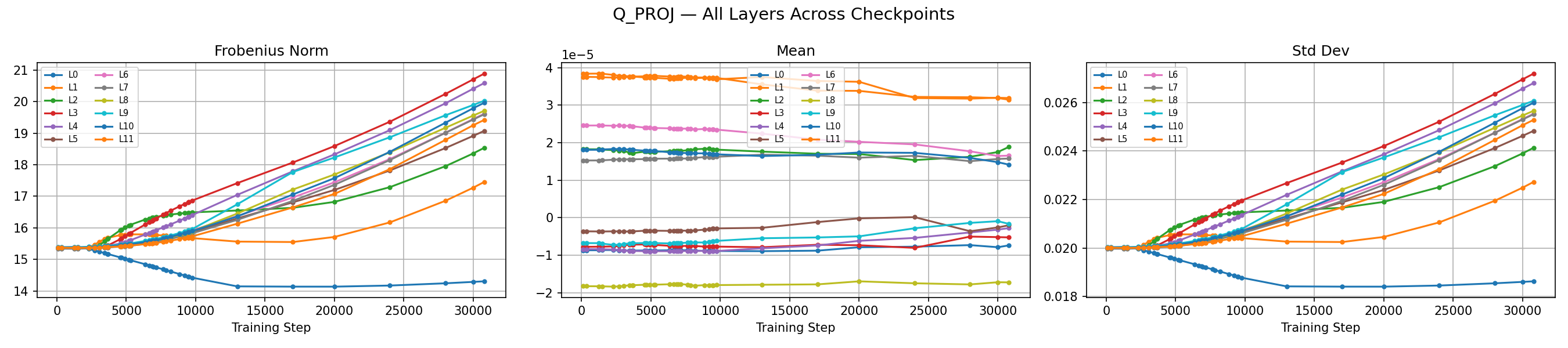}
    \caption{Q (Query) matrix analysis across different heads}
    \label{fig:Q}
\end{figure*}

\begin{figure*}[t]
    \centering
    \includegraphics[width=\textwidth]{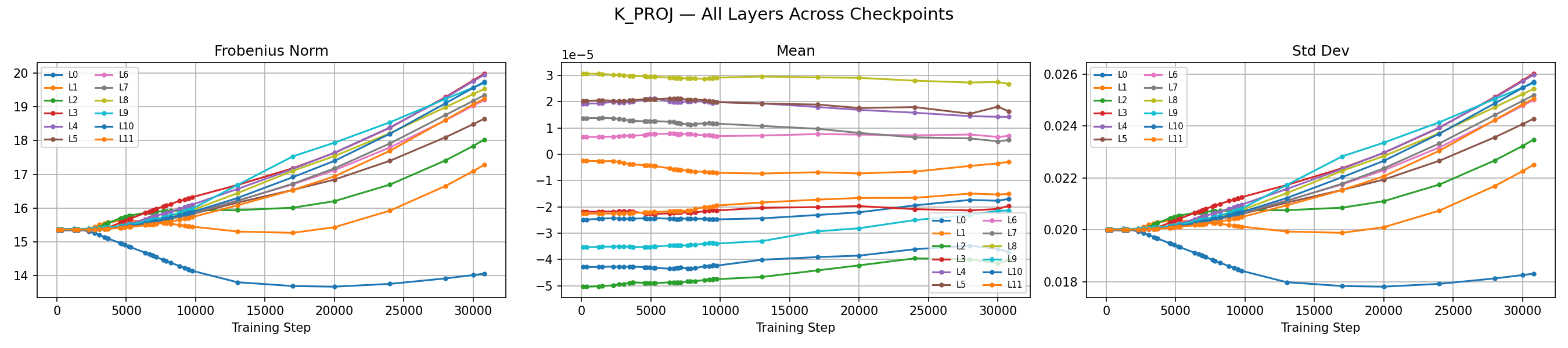}
    \caption{K (Key) matrix analysis across different heads}
    \label{fig:K}
\end{figure*}

\begin{figure*}[t]
    \centering
    \includegraphics[width=\textwidth]{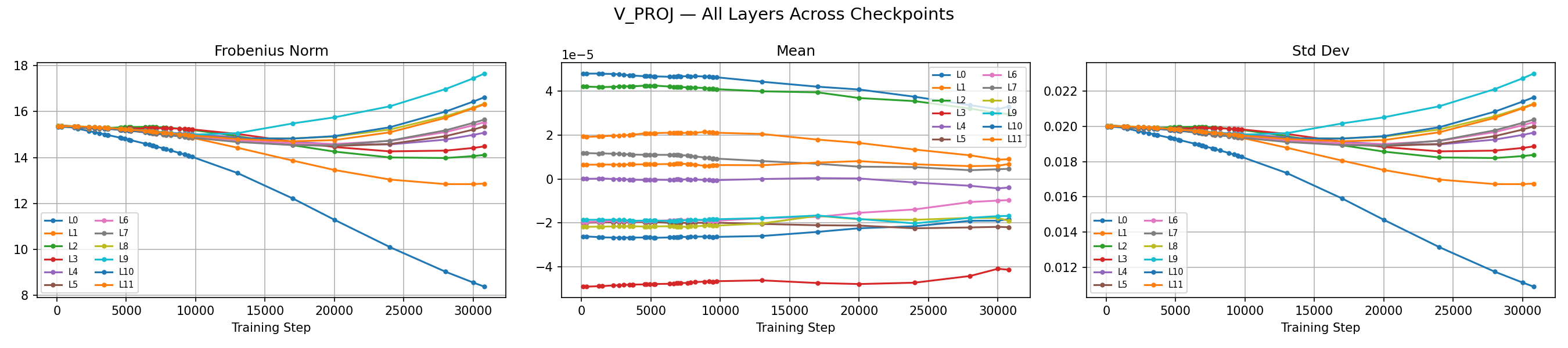}
    \caption{V (Vector) matrix analysis across different heads}
    \label{fig:V}
\end{figure*}

\begin{figure*}[t]
    \centering
    \includegraphics[width=\textwidth]{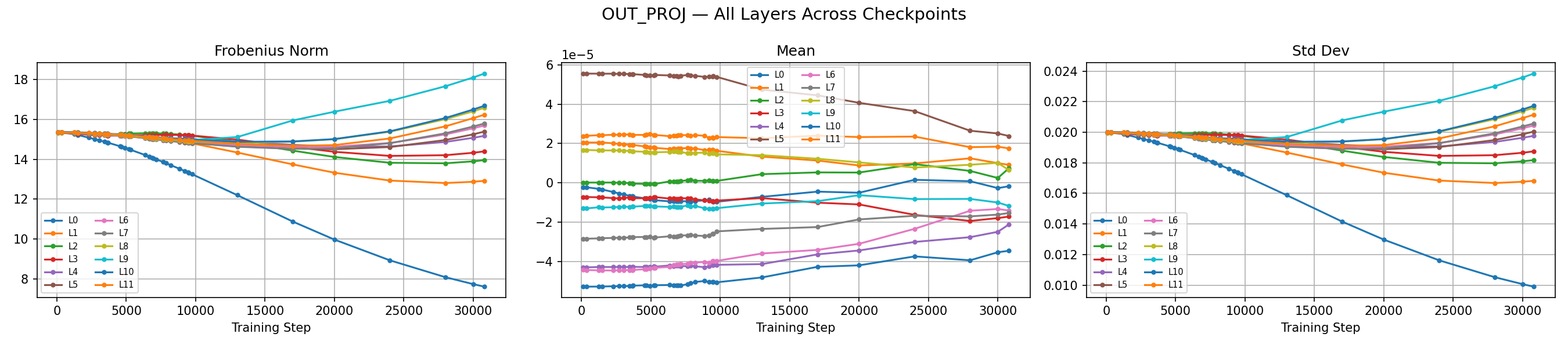}
    \caption{O (output) matrix analysis across different heads}
    \label{fig:O}
\end{figure*}

Across the 12 transformer layers (Figures~\ref{fig:Q},\ref{fig:K}, \ref{fig:V}, and \ref{fig:O}),
we observe that early training checkpoints (<4000) show nearly identical statistics across layers, indicating that the projections are
initially similar due to common initialization.
Around training steps \(4{,}000\!-\!6{,}000\),
the Frobenius norms and standard deviations begin to diverge across layers, suggesting the onset of layer specialization. Later layers exhibit stronger growth in parameter magnitude,
particularly for the query and key projections,
while some lower layers remain relatively stable. These observations coincide with the findings of change point detection across different blimp categories.

This divergence is consistent with the functional role of multi-head attention: different heads and layers learn to focus on distinct aspects of the input sequence. For example, some heads specialize in capturing syntactic dependencies (e.g., subject–verb relations), others attend to positional locality between nearby tokens, and others capture semantic similarity between related concepts.
The increasing separation of parameter norms across layers therefore reflects the progressive specialization of attention mechanisms as training proceeds.

\begin{figure*}[t]
    \centering
    \includegraphics[width=\textwidth]{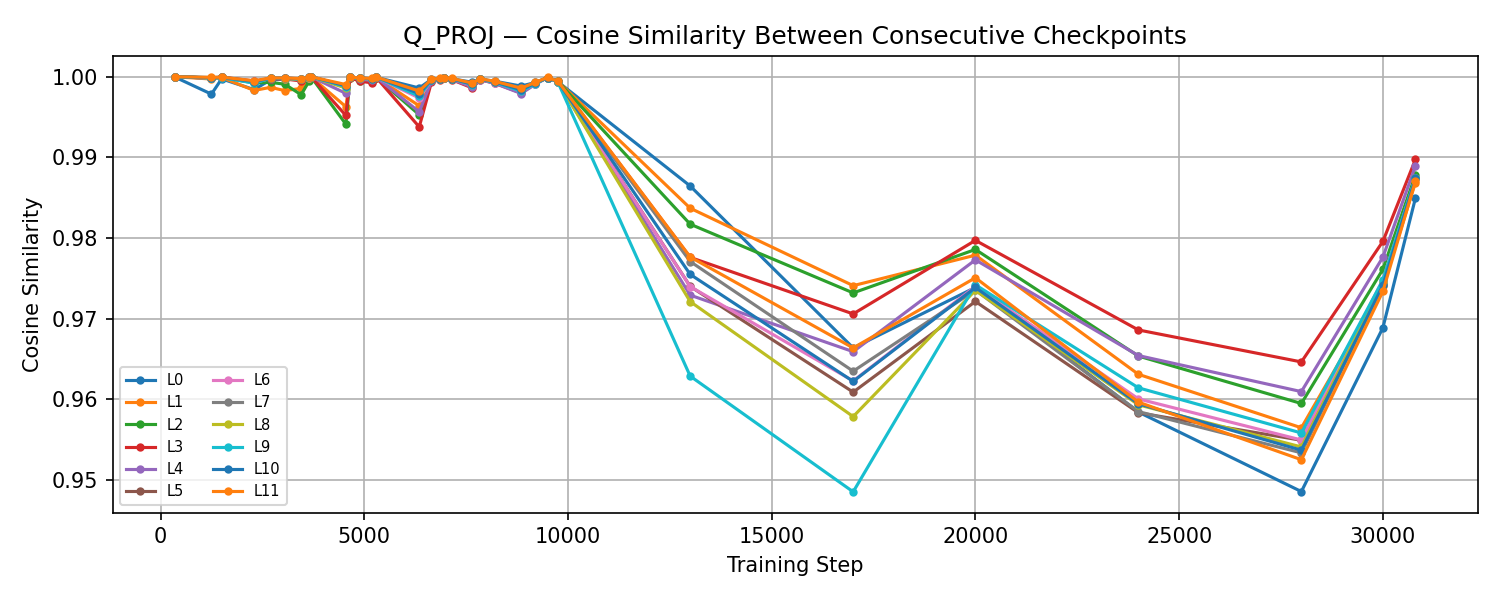}
    \caption{Q (Query) matrix cosine similarity analysis across different heads}
    \label{fig:Q_sim}
\end{figure*}

\begin{figure*}[t]
    \centering
    \includegraphics[width=\textwidth]{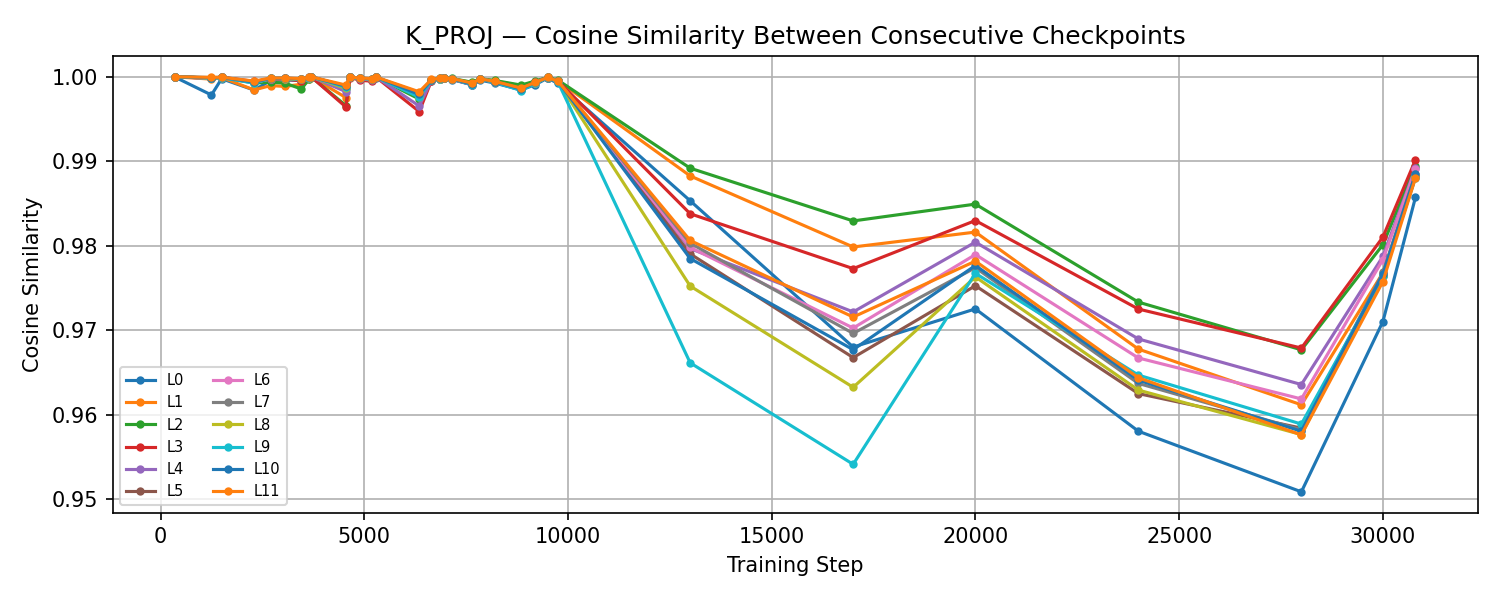}
    \caption{K (Key) matrix cosine similarity analysis across different heads}
    \label{fig:K_sim}
\end{figure*}

\begin{figure*}[t]
    \centering
    \includegraphics[width=\textwidth]{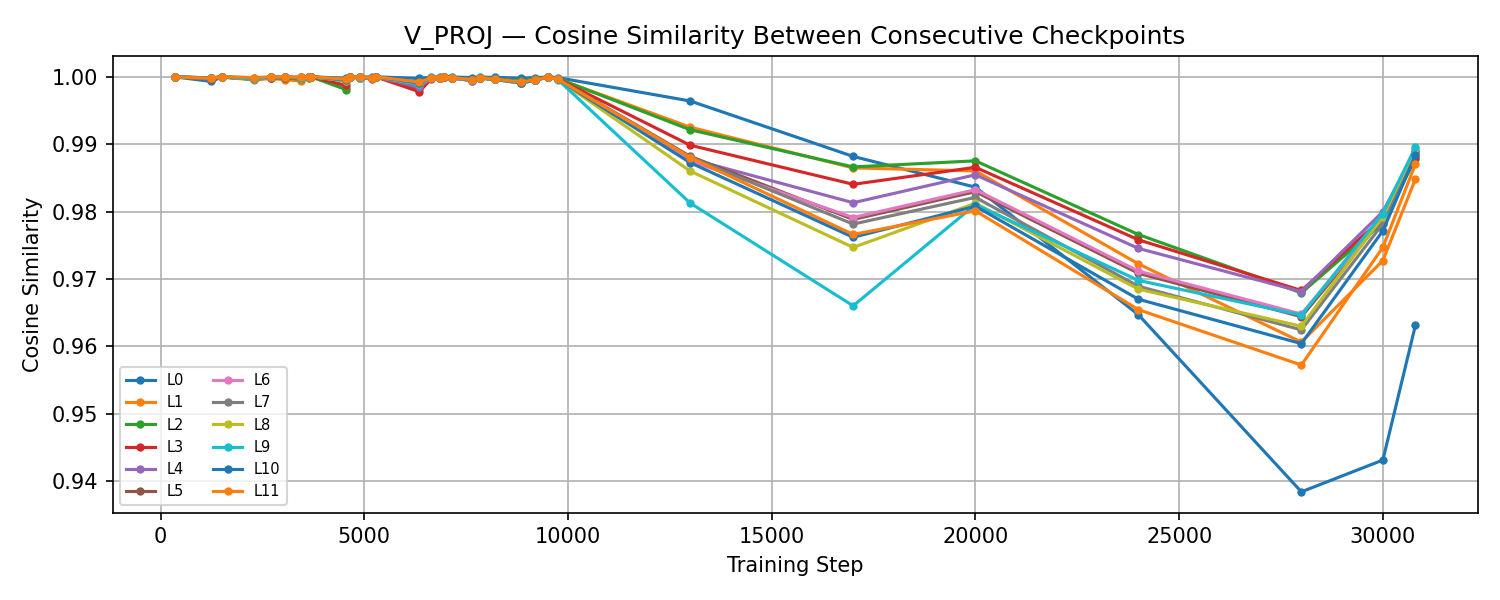}
    \caption{V (Vector) matrix cosine similarity analysis across different heads}
    \label{fig:V_sim}
\end{figure*}

To further analyze how attention parameters evolve during training, we measure the similarity of projection matrices between consecutive training checkpoints (Figures~\ref{fig:Q_sim},\ref{fig:K_sim}, and \ref{fig:V_sim}).
Let \(W^{(t)}\) denote a projection matrix
(\(W_Q, W_K, W_V\), or \(W_O\)) at training step \(t\).
We compute the cosine similarity between flattened parameter vectors
at two successive checkpoints:

\[
\mathrm{cos}(W^{(t)}, W^{(t-1)})
=
\frac{\langle w^{(t)}, w^{(t-1)} \rangle}
{\|w^{(t)}\|_2 \, \|w^{(t-1)}\|_2},
\]

where

\[
w^{(t)} = \mathrm{vec}(W^{(t)})
\]

denotes the vector obtained by flattening the matrix \(W^{(t)}\)
and \(\langle \cdot,\cdot \rangle\) denotes the Euclidean inner product.
Cosine similarity values close to \(1\) indicate that parameters
change very little between checkpoints, while lower values indicate
larger updates during training.

For each decoder layer \(l=1,\dots,12\) and each projection matrix
\(W\in\{W_Q,W_K,W_V,W_O\}\), we compute

\[
\mathrm{cos}\big(W_l^{(t)},W_l^{(t-1)}\big)
\]

across the sequence of training checkpoints.
This provides a layer-wise trajectory of how rapidly parameters evolve
during training. The range is .94 to 1. 

The cosine similarity curves reveal distinct training phases.
During the earliest checkpoints (<10,000), similarities are extremely close to
\(1\), indicating that updates are small and parameters remain close to
their initialization. As training progresses, similarities begin to decrease and diverge across layers, reflecting larger updates and the emergence of layer-specific specialization.
Later in training, cosine similarities increase again, suggesting that parameter updates stabilize as the model approaches convergence.

\begin{figure*}[t]
    \centering
    \includegraphics[width=\textwidth]{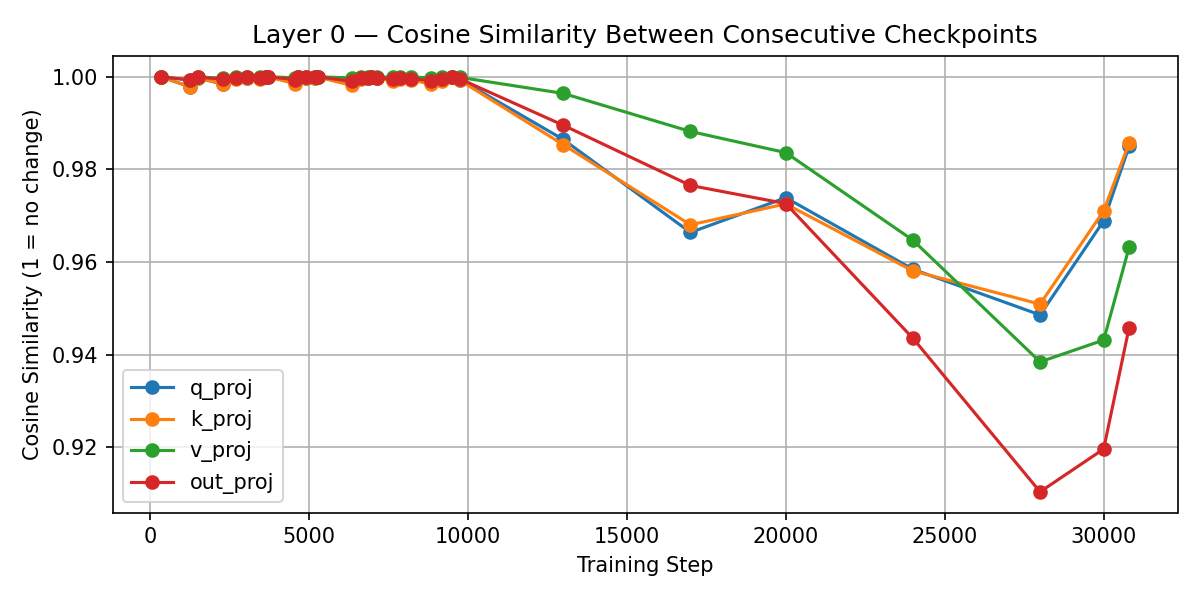}
    \caption{Q,K,V,O matrix cosine similarity analysis across the first layer}
    \label{fig:1st_layer_sim}
\end{figure*}

For a specific layer, the four different similarity matrices are provided in Figure~\ref{fig:1st_layer_sim}.

To understand if some of the layers are failing in the training process, we conducted several experiments.

For a projection matrix $W \in \mathbb{R}^{d_{\mathrm{out}} \times d_{\mathrm{in}}}$
with singular value decomposition $W = U\Sigma V^\top$, let
\[
    \sigma_1 \geq \sigma_2 \geq \cdots \geq \sigma_r > 0,
    \qquad r = \min(d_{\mathrm{out}},\, d_{\mathrm{in}}),
\]
denote the singular values in descending order. We track the following
four spectral quantities at each checkpoint.

\paragraph{Largest and smallest singular values.}
\[
    \sigma_{\max} = \sigma_1,
    \qquad
    \sigma_{\min} = \sigma_r.
\]
A sustained decay $\sigma_{\min} \to 0$ would indicate that $W$ is
approaching rank deficiency and the corresponding layer is ceasing to
contribute independent information to the residual stream.

\paragraph{Condition number.}
\[
    \kappa(W) = \frac{\sigma_{\max}}{\sigma_{\min}}.
\]
A well-conditioned matrix has $\kappa(W)$ close to unity; as
$\kappa(W) \to \infty$ the matrix becomes singular. We add a small
stabilising constant $\varepsilon = 10^{-12}$ in practice to avoid
division by zero.

\paragraph{Effective rank.}
Rather than counting all non-zero singular values, which is sensitive
to numerical noise, we define the effective rank as the minimum number
of singular values required to capture a fraction $\tau = 0.99$ of the
total spectral energy:
\[
    \mathrm{rank}_{\mathrm{eff}}(W;\tau)
    \;=\;
    \min\!\left\{
        k \in \{1,\dots,r\}
        \;\middle|\;
        \frac{\displaystyle\sum_{i=1}^{k} \sigma_i^2}
             {\displaystyle\sum_{i=1}^{r} \sigma_i^2}
        \;\geq\; \tau
    \right\}.
\]
Note that the denominator equals $\|W\|_F^2$, so the criterion is
equivalent to asking which $k$ directions account for at least
$100\tau$\% of the Frobenius energy. A monotonically decreasing
$\mathrm{rank}_{\mathrm{eff}}$ over training reflects normal
representational compression; a collapse toward unity would signal
pathological rank reduction.

\begin{figure}[htbp]
    \centering
    \includegraphics[width=\textwidth]{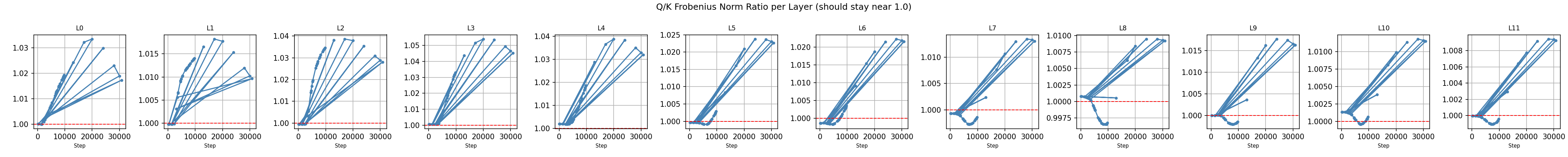}
    \caption{Ratio $\|W_Q\|_F / \|W_K\|_F$ per layer across training.
             All layers remain within 5\% of unity throughout, indicating
             well-balanced attention logit scaling.}
    \label{fig:qk_ratio}
\end{figure}

\begin{figure}[htbp]
    \centering
    \includegraphics[width=\textwidth]{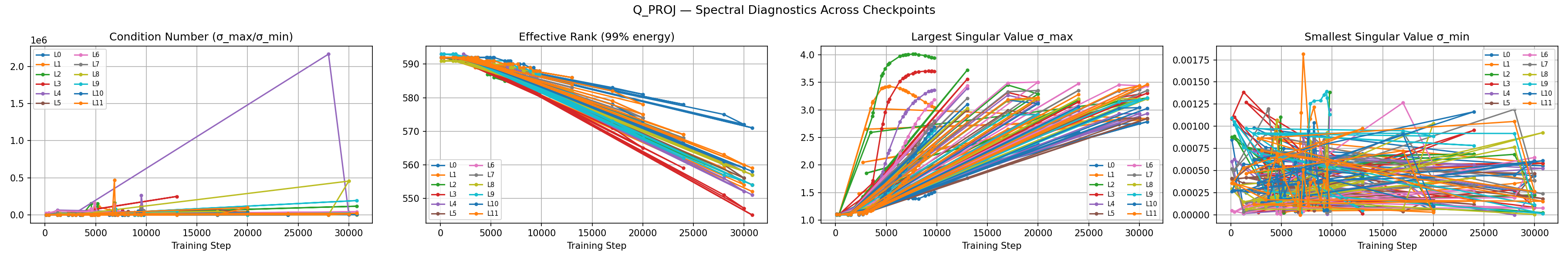}
    \caption{Spectral diagnostics for $W_Q$: condition number
             $\sigma_{\max}/\sigma_{\min}$, effective rank (99\% energy
             threshold), largest singular value $\sigma_{\max}$, and
             smallest singular value $\sigma_{\min}$.}
    \label{fig:q_spectral}
\end{figure}

\begin{figure}[htbp]
    \centering
    \includegraphics[width=\textwidth]{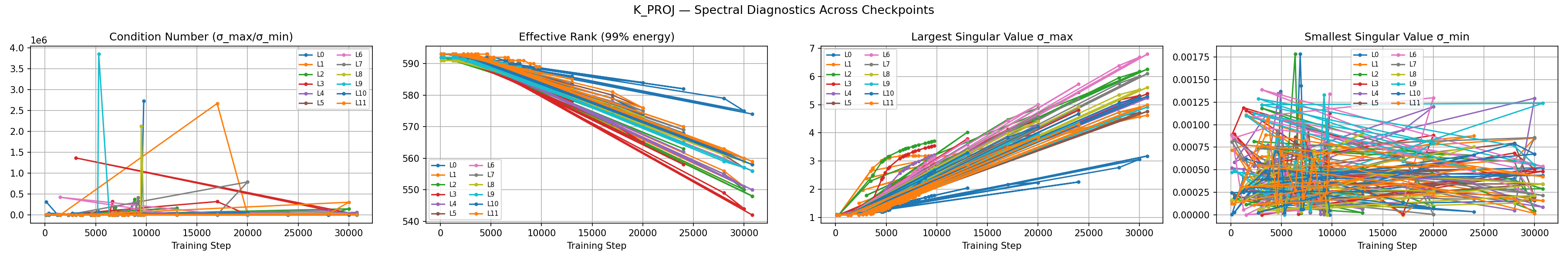}
    \caption{Spectral diagnostics for $W_K$.}
    \label{fig:k_spectral}
\end{figure}

\begin{figure}[htbp]
    \centering
    \includegraphics[width=\textwidth]{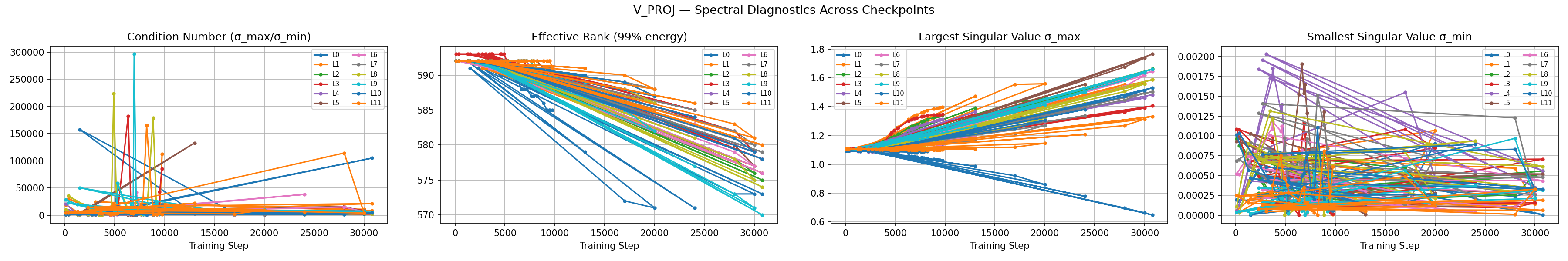}
    \caption{Spectral diagnostics for $W_V$.}
    \label{fig:v_spectral}
\end{figure}

\begin{figure}[htbp]
    \centering
    \includegraphics[width=\textwidth]{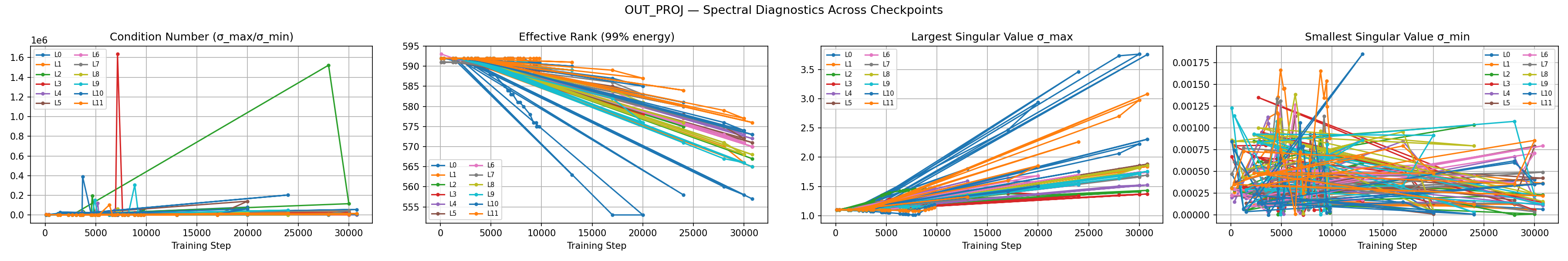}
    \caption{Spectral diagnostics for $W_O$.}
    \label{fig:out_spectral}
\end{figure}

We examined the evolution of the four attention projection matrices
$W_Q$, $W_K$, $W_V$, and $W_O$ across all 12 layers of a causal
transformer over 30{,}000 training steps, tracking the Frobenius norm,
entry-wise mean and standard deviation, the Q/K norm ratio, and a full
suite of spectral diagnostics, including condition number, effective rank,
and the largest and smallest singular values.

\paragraph{Q/K norm balance.}
Figure~\ref{fig:qk_ratio} plots the ratio $\|W_Q\|_F / \|W_K\|_F$ for
each layer throughout training. Across all 12 layers, this ratio remains
within 5\% of unity at every checkpoint. This is a direct empirical test
of the ill-conditioning mechanism described by \citet{agarwal2026}: when
the Q and K norms diverge substantially, attention logit magnitudes
become imbalanced and the softmax distribution collapses toward a
one-hot or uniform extreme. The observed near-unity ratio confirms that
no such imbalance develops, and that the attention mechanism remains
well-scaled throughout training.

\paragraph{Spectral diagnostics.}
Figures~\ref{fig:q_spectral}--\ref{fig:out_spectral} present the spectral
diagnostics for each projection. Three observations are noteworthy.
First, the condition number $\sigma_{\max}/\sigma_{\min}$ exhibits large
transient spikes in early training (steps 0--10{,}000), reaching values
of order $10^5$--$10^6$ in some layers, but stabilises to low values
by step 15{,}000 and remains stable thereafter. Such early training
transients are a well-documented consequence of the optimizer exploring
a high-curvature loss landscape before settling into a stable regime,
and their subsequent resolution indicates no persistent ill-conditioning.
Second, the effective rank (defined as the number of singular values
required to capture 99\% of the total spectral energy) decreases
monotonically from approximately 590--592 at initialization to
550--575 by step 30{,}000, uniformly across all layers and projections.
This progressive rank compression is the expected signature of
representational specialization: the network learns to express its
transformations in a lower-dimensional subspace as training proceeds.
Critically, the effective rank remains well above unity throughout,
ruling out rank collapse or functional layer death.
Third, and most directly relevant to the health of individual layers,
the smallest singular value $\sigma_{\min}$ does not collapse to zero
in any layer or projection over the course of training. A sustained
$\sigma_{\min} \to 0$ trajectory would indicate that the weight matrix
is becoming singular and the layer is ceasing to contribute independent
information to the residual stream; no such trajectory is observed.

\paragraph{Overall assessment.}
Taken together, the Frobenius norm trajectories, the Q/K balance
analysis, and the spectral diagnostics consistently indicate that all
12 attention layers remain well-conditioned and functionally active
throughout the 30{,}000-step training run. The model does not exhibit
the spectral signatures of attention collapse, layer death, or
proximity to the ill-conditioned barrier identified by
\citet{agarwal2026}. Training is proceeding healthily.

\end{document}